\documentclass[10pt,twocolumn,letterpaper]{article}

\usepackage{cvpr}              

\usepackage{graphicx}
\usepackage{amsmath}
\usepackage{amssymb}
\usepackage{booktabs}

\usepackage[pagebackref,breaklinks,colorlinks]{hyperref}

\usepackage[capitalize]{cleveref}
\crefname{section}{Sec.}{Secs.}
\Crefname{section}{Section}{Sections}
\Crefname{table}{Table}{Tables}
\crefname{table}{Tab.}{Tabs.}

\usepackage{multirow}
\usepackage{amsmath,bm}
\usepackage{mathrsfs}
\usepackage{overpic}
\usepackage{booktabs}
\usepackage{pifont}
\usepackage{algorithm}
\usepackage{algorithmic}
\usepackage{balance}
\graphicspath{{./Imgs/}}

\usepackage{colortbl}
\usepackage{color,xcolor}

\def\cvprPaperID{9854} 
\def\confName{CVPR}
\def\confYear{2023}

\begin{document}

\title{Feature Shrinkage Pyramid for Camouflaged Object Detection \\ with Transformers}

\author{
Zhou Huang$^{1,2}$\protect\footnotemark[2]\, \quad
Hang Dai$^3$\protect\footnotemark[2]\, \quad
Tian-Zhu Xiang$^4$\protect\footnotemark[1]\, \quad
Shuo Wang$^5$\, \\
Huai-Xin Chen$^2$\, \quad
Jie Qin$^6$\, \quad
Huan Xiong$^{7}$\\
$^1$Sichuan Changhong Electric Co., Ltd., China\quad
$^2$UESTC, China \quad $^3$University of Glasgow, UK \\
$^4$G42, UAE \quad
$^5$ETH Zurich, Switzerland \quad
$^6$CCST, NUAA, China \quad
$^7$MBZUAI, UAE \\
{\tt\small 
chowhuang@std.uestc.edu.cn,
hang.dai@glasgow.ac.uk,
\{tianzhu.xiang19, }\\
\tt\small {
shawnwang.tech, qinjiebuaa, huan.xiong.math\}@gmail.com, 
huaixinchen@uestc.edu.cn}
}

\maketitle

{
\renewcommand{\thefootnote}{\fnsymbol{footnote}} 
\footnotetext[2]{Equal contributions. *Corresponding author:~\textit{Tian-Zhu Xiang}.}
\renewcommand{\thefootnote}{\arabic{footnote}}
}

\begin{abstract}
\vspace{-10pt}
Vision transformers have recently shown strong global context modeling capabilities in camouflaged object detection. However, they suffer from two major limitations: less effective locality modeling and insufficient feature aggregation in decoders, which are not conducive to camouflaged object detection that explores subtle cues from indistinguishable backgrounds. To address these issues, in this paper, we propose a novel transformer-based Feature Shrinkage Pyramid Network (FSPNet), which aims to hierarchically decode locality-enhanced neighboring transformer features through progressive shrinking for camouflaged object detection. 
Specifically, we propose a non-local token enhancement module (NL-TEM) that employs the non-local mechanism to interact neighboring tokens and explore graph-based high-order relations within tokens to enhance local representations of transformers. 
Moreover, we design a feature shrinkage decoder (FSD) with adjacent interaction modules (AIM), which progressively aggregates adjacent transformer features through a layer-by-layer shrinkage pyramid to accumulate imperceptible but effective cues as much as possible for object information decoding. 
Extensive quantitative and qualitative experiments demonstrate that the proposed model significantly outperforms the existing 24 competitors on three challenging COD benchmark datasets under six widely-used evaluation metrics. Our code is publicly available at \url{https://github.com/ZhouHuang23/FSPNet}. 
\end{abstract}

\section{Introduction}
\label{sec:intro}

Camouflage is a common defense or tactic in organisms that ``perfectly'' blend in with their surroundings to deceive predators (prey) or sneak up on prey (hunters).  
Camouflaged object detection (COD)~\cite{fan2020camouflaged} aims to segment camouflaged objects in the scene and has been widely applied in species conservation~\cite{nafus2015hiding}, medical image segmentation~\cite{chen2022camouflaged,li2022trich}, and industrial defect detection~\cite{bhajantri2006camouflage}, etc.

\begin{figure}[t]
	\centering
	\begin{overpic}[width=0.95\linewidth]{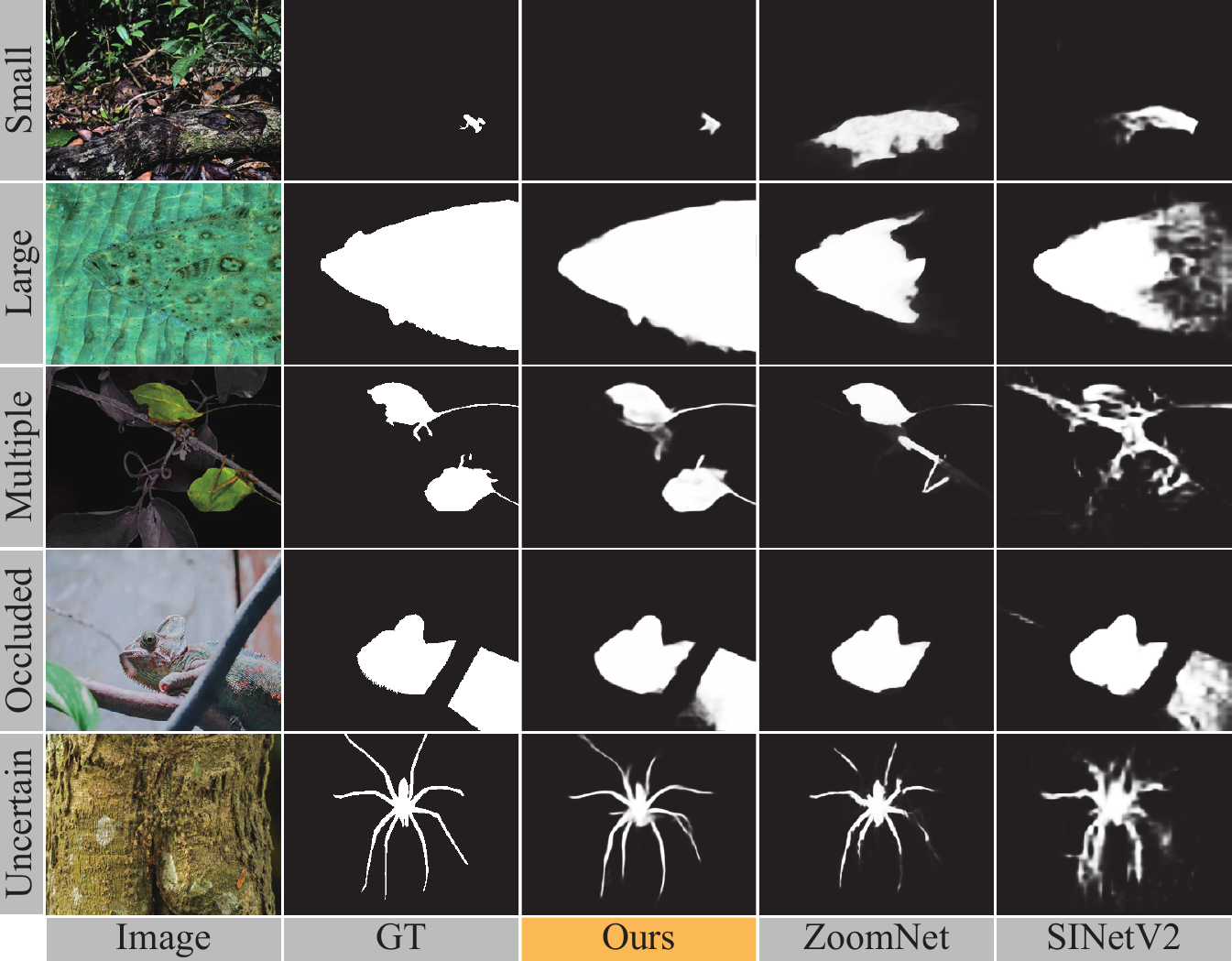}
	\end{overpic}
	\caption{\textbf{Visual comparison of COD in different challenging scenarios}, including small, large, multiple, occluded and boundary-uncertain camouflaged objects. Compared with the recently proposed \textit{ZoomNet}~\cite{youwei2022zoom} and \textit{SINet-v2}~\cite{fan2021concealed}, our method provides superior performance with more accurate object localization and more complete object segmentation, mainly due to the proposed locality-enhanced global context exploration and progressive shrinkage decoder.
	}
	\label{fig:First-VC}
\end{figure}

Due to the high similarities between camouflaged objects and their backgrounds, camouflaged objects are usually inconspicuous and indistinguishable, which brings great challenges to accurate detection. 
Recently, the development of deep learning and the availability of large-scale COD datasets (\textit{e.g.}, COD10K~\cite{fan2020camouflaged}) have significantly advanced camouflaged object detection. Numerous deep learning-based methods have been proposed, which can be roughly divided into three categories: targeted design of feature exploration modules, multi-task joint learning frameworks, and bio-inspired methods. 
Although these methods have made remarkable progress, they mainly rely heavily on convolutional neural networks (CNNs), which cannot capture long-range dependencies due to the limited receptive fields, resulting in inferior performance for COD. 
As shown in Fig.~\ref{fig:First-VC}, recently proposed state-of-the-art CNN-based methods (\textit{e.g.}, ZoomNet~\cite{youwei2022zoom} and SINet-v2~\cite{fan2021concealed}) fail to explore global feature relations and thus often provide predictions of incomplete object regions, especially for multiple objects, large objects and occlusion cases.
Although larger convolution kernels or simply stacking multiple convolution layers with small kernels can enlarge receptive fields and thus alleviate this issue to some extent, it also dramatically increases the computational cost and the number of network parameters. Furthermore, studies~\cite{shi2021video} have shown that simply network deepening is ineffective for long-range dependency modeling.

Compared to CNNs, vision transformers (ViT)~\cite{dosovitskiy2020image}, which have recently been introduced into computer vision and demonstrated significant breakthroughs in various vision applications~\cite{khan2021transformers}, can efficiently model long-range dependencies with the self-attention operations and thus overcome the above drawbacks of CNNs-based models.
Recently, the works of \cite{yang2021uncertainty} and \cite{liu2022boost} have attempted to accommodate transformers for COD and shown promising performance. These methods either employ transformer as a network component for feature decoding or utilize the off-the-shelf vision transformers as backbones for feature encoding. 
Through a thorough analysis of these methods for COD, we observe two major issues within existing techniques: 
1) \textit{Less effective local feature modeling for transformer backbones.} We argue that both global context and local features play essential roles in COD tasks. However, we observe that most transformer-based methods lack a locality mechanism for information exchange within local regions. 
2) \textit{Limitations of feature aggregation in decoders.}  
Existing decoders (shown in Fig.~\ref{fig:DS} (a)-(d)) usually directly aggregate the features with significant information differences (\textit{e.g.}, low-level features with rich details and high-level features with semantics), which tends to discard some inconspicuous but valuable cues or introduce noise, resulting in inaccurate predictions. This is a big blow for the task of identifying camouflaged objects from faint clues.

To this end, in this paper, we propose a novel transformer-based Feature Shrinkage Pyramid Network, named \textit{FSPNet}, which aims to hierarchically decode neighboring transformer features which are locality-enhanced global representations for camouflaged objects through progressive shrinking, thereby excavating and accumulating rich local cues and global context of camouflaged objects in our encoder and decoder for accurate and complete camouflaged object segmentation.
Specifically, to complement local feature modeling in the transformer encoder, we propose a non-local token enhancement module (NL-TEM) which employs the non-local mechanism to interact neighboring similar tokens and explore graph-based high-level relations within tokens to enhance local representations. 
Furthermore, we design a feature shrinkage decoder (FSD) with adjacent interaction modules (AIMs) which progressively aggregates adjacent transformer features in pairs through a layer-by-layer shrinkage pyramid architecture to accumulate subtle but effective details and semantics as much as possible for object information decoding. 
Owing to the global context modeling of transformers, locality exploration within tokens and progressive feature shrinkage decoder, our proposed model achieves state-of-the-art performance and provides an accurate and complete camouflaged object segmentation. Our main contributions are summarized as follows:
\begin{itemize}
    \item We propose a non-local token enhancement module (NL-TEM) for feature interaction and exploration between and within tokens to compensate for locality modeling of transformers. 
    
    \item We design a feature shrinkage decoder (FSD) with the adjacent interaction module (AIM) to better aggregate camouflaged object cues between neighboring transformer features through progressive shrinking for camouflaged object prediction.
    
    \item Comprehensive experiments show that our proposed FSPNet achieves superior performance on three widely-used COD benchmark datasets compared to 24 existing state-of-the-art methods.  
    
\end{itemize}

\section{Related Work}
\label{relatedwork}
\subsection{CNN-based Camouflaged Object Detection}
Recently, CNN-based approaches have made impressive progress on the COD task by releasing large-scale datasets. 
Some works attempt to mine inconspicuous features of camouflage objects from the background through meticulously designed feature exploration modules,\textit{ e.g.}, contextual feature learning~\cite{mei2021camouflaged,sun2021context}, texture-aware learning~\cite{zhu2021inferring}, and frequency-domain learning~\cite{zhong2022detecting}. 
There are also some models~\cite{li2021uncertainty,yang2021uncertainty,liu2022modeling} which propose to model uncertainty in data labeling or camouflaged data itself for COD. 
Besides, the multi-task learning framework is commonly used for COD. These methods generally introduce auxiliary tasks such as classification~\cite{le2019anabranch}, edge/boundary detection~\cite{zhai2021mutual,zhu2022can,sun2022Boundary}, and object ranking~\cite{lv2021simultaneously}.
Furthermore, some methods detect camouflaged objects by mimicking behavior patterns or visual mechanics of predators such as the search and identification process~\cite{fan2021concealed}, and zooming in and out~\cite{youwei2022zoom,jia2022segment}. 
In addition to image-based COD, more recently, \cite{cheng2022implicit} proposed to discover camouflaged objects in videos using motion information.
Although CNN-based models have achieved promising performance, these methods do not explore long-range dependencies due to limited receptive fields, which is critical for COD in images containing diverse objects.

\subsection{Decoding Strategy}
\begin{figure}[t]
	\centering
	\begin{overpic}[width=1\linewidth]{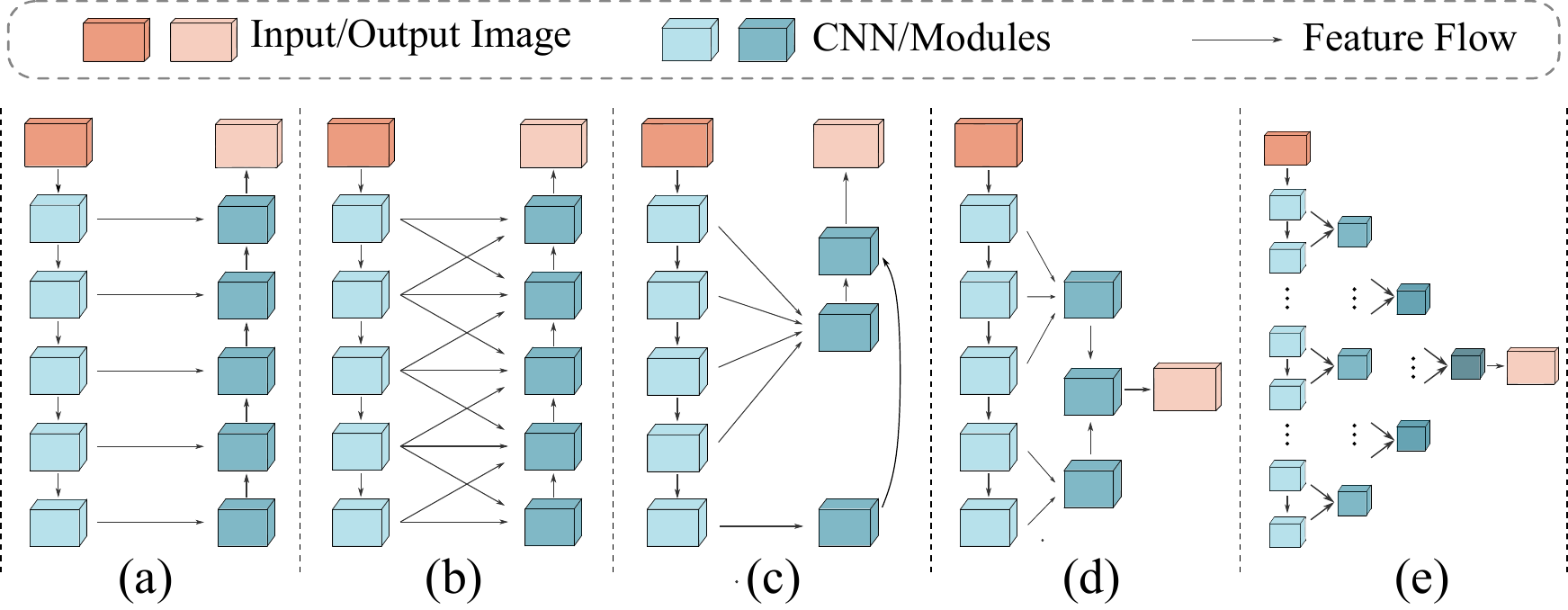}
	\end{overpic}
	\caption{\textbf{Different types of decoding structures for  object segmentation.} (a) U-shaped decoding structure \cite{zhou2020interactive,youwei2022zoom,siris2021scene}. (b) Dense integration strategy \cite{zhang2017amulet,zhang2018deep,pang2020multi}. (c) Feedback refinement strategy~\cite{zhao2020suppress,zhai2021mutual,zhu2022can}. (d) Separate decoding of low-level and high-level features \cite{huang2021semantic,fan2020camouflaged,fan2020pranet}. (e) Our decoding structure.}
	\label{fig:DS} 
\end{figure}

By reviewing vision tasks related to COD (\emph{e.g.}, salient object detection and medical image segmentation), the decoder design can be summarized into four typical feature decoding strategies: (a) U-shaped decoding structure, (b) dense integration strategy, (c) feedback refinement strategy, and (d) separate decoding of low-level and high-level features, as shown in Fig.~\ref{fig:DS}. 
Specifically, as the most prevalent feature decoding strategy, the U-shaped decoders~\cite{zhou2020interactive,youwei2022zoom,siris2021scene} integrate lateral output multi-scale backbone features and recover object details gradually in a bottom-up manner. To weaken the interference of large resolution differences on the compatibility of feature fusion, some methods \cite{zhang2017amulet,zhang2018deep,pang2020multi} use a dense integration strategy to aggregate multi-level features. Some methods treat the high-level output features separately (usually the last layer of backbone features) and then integrate them with other lateral outputs to improve localization and segmentation results. Some other methods~\cite{huang2021semantic,fan2020camouflaged,fan2020pranet} deal with low-level and high-level features differently to explore and integrate local cues and global semantics for object segmentation. Unlike the mainstream decoding strategy mentioned above, we adopt a pyramidal shrinkage decoding strategy, as shown in Fig.~\ref{fig:DS} (e), which aggregates adjacent features and recovers the object information layer by layer in a progressive manner.

\subsection{Vision Transformer}

Transformers, which are initially designed for natural language processing~\cite{vaswani2017attention}, have been widely applied in computer vision in recent years and achieved significant progress in numerous visual applications, such as image classification \cite{dosovitskiy2020image}, object detection \cite{carion2020end}, and semantic segmentation \cite{zheng2021rethinking}. 
Benefiting from the self-attention mechanism, transformers are better at capturing long-range dependencies when compared to CNN-based models~\cite{wang2018non,tay2021synthesizer}.
To our knowledge, ViT~\cite{dosovitskiy2020image} is the first transformer model in the computer vision community, which directly takes sequences of image patches as input to explore long-range spatial correlations for the classification task. Then a series of improved versions sprung up, such as data-efficient image transformers (DeiT)~\cite{touvron2021training}, pyramid vision transformer~\cite{wang2021pyramid}, and Swin transformer~\cite{liu2021swin}. 
For camouflaged object detection, \cite{pei2022osformer} propose a one-stage transformer framework for camouflaged instance segmentation. 
\cite{yang2021uncertainty}, \cite{liu2022boost}, and \cite{hu2022high} have made some attempts to detect camouflaged objects using transformers and achieved good performance. 
However, these methods remain limitations in the exploration of locality modeling and feature aggregation of decoders inherited from the CNN design paradigm, \textit{i.e.}, information loss caused by large-span aggregation. 
In this paper, we design a feature shrinkage decoder with the adjacent interaction module to progressively aggregate adjacent features through the shrinkage pyramid for accurate decoding.

\begin{figure*}[htb]
	\centering
	\begin{overpic}[width=0.9\linewidth]{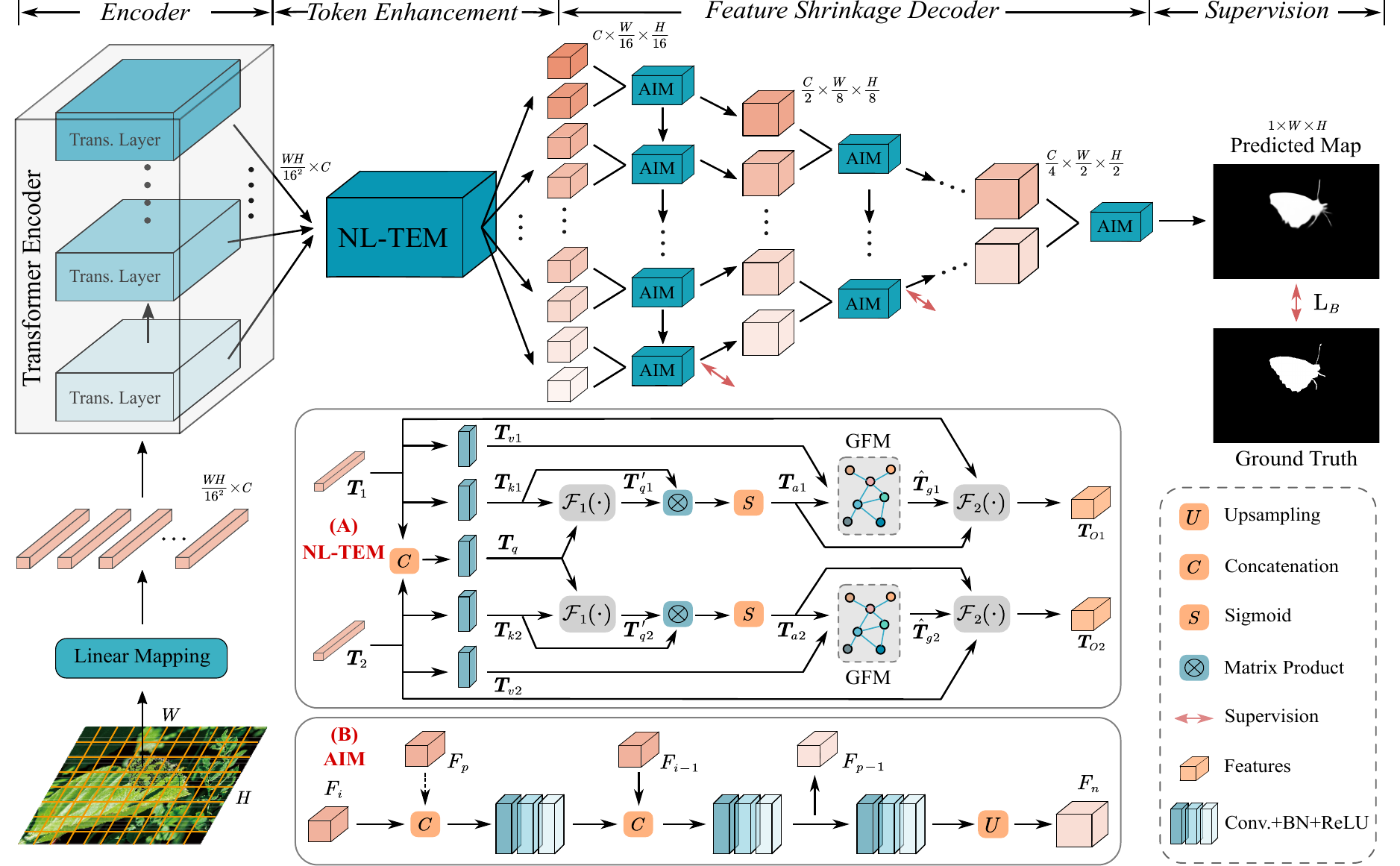}
	\end{overpic}
	\caption{\textbf{Overall architecture of the proposed FSPNet.} It consists of three key components: a ViT-based encoder, a non-local token enhancement module (NL-TEM) and a feature shrinkage decoder (FSD) with adjacent interaction modules (AIM).
	} 
	\label{fig:Flow}
\end{figure*}

\section{Proposed Method} 
\label{Proposed Method}

\subsection{Overview}
\cref{fig:Flow} illustrates the overall architecture of our proposed FSPNet model. The main components include a vision transformer encoder, a non-local token enhancement module (NL-TEM), and a feature shrinkage decoder (FSD). 
Specifically, the input image is first serialized into tokens as input to a transformer encoder to model global contexts using the self-attentive mechanism. 
After that, to strengthen the local feature representation within tokens, a non-local token enhancement module (NL-TEM) is designed to perform feature interaction and exploration between and within tokens and convert the enhanced tokens from the encoder space to the decoder space for decoding.
In the decoder, to merge and retain subtle but critical cues as much as possible, we design a feature shrinkage decoder (FSD) to progressively aggregates adjacent features through layer-by-layer shrinkage to decode object information.

\subsection{Transformer Encoder}
Unlike previous works on COD, we utilize a vanilla vision transformer (ViT) as the encoder to model the global context of camouflaged objects, mainly consisting of image serialization and transformer layers. a) Serialization. In order to satisfy the self-attention requirement on the input and reduce the computational complexity, inspired by \cite{dosovitskiy2020image}, the given image $\boldsymbol{I}\in \mathbb{R} ^{C\times H\times W} $ is first split into a sequence of non-overlapping image patches with patch size ($s,s$), where $C$, $H$ and $W$ denote channel size, height and width of image $\boldsymbol{I}$, respectively, and $s=16$ in our experiments.  
Then the image patches are linearly projected into a 1D sequence of token embeddings $T^0\in \mathbb{R} ^{l\times d}$, where $l={HW}/{s^2}$ is the sequence length and $d=s^2\cdot C$ is the embedding dimension. 
b) Transformer layer. To preserve positional information, an additional learnable position embedding $E^p$ is added to the tokens, forming the new tokens $T^p=T^0+E^p$. 
All the tokens are then input into a transformer encoder with $n$ transformer layers, where each layer contains a multi-head self-attention (MSA) and a multi-layer perceptron (MLP) block. It can be formulated as:  
\begin{equation}
    \boldsymbol{T}={\rm{MLP}}({\rm{MSA}}(\boldsymbol{T}^{p})),
\end{equation}
where $\boldsymbol{T}\in\mathbb{R}^{l\times c} $, $c$ is the token dimension. 
Note that layer normalization~\cite{ba2016layer} is applied before each block and residual connections after each block. Thus, we obtain the output tokens from the encoder.

\subsection{Non-local Token Enhancement Module}
Transformers bring powerful global context modeling capabilities but lack a locality mechanism for information exchange within a local region. 
Besides, it is well known that camouflaged targets always share very similar appearance information with noise objects and background, where the slight differences are difficult to be distinguished by low-order relations. 
Inspired by~\cite{te2020edge,wang2018non}, we design a non-local token enhancement module (NL-TEM) which is applied on neighboring tokens (local region) to strengthen the local feature representation. A non-local operation is first adopted to interact adjacent similar tokens for aggregation of adjacent camouflage clues. Then a graph convolution network (GCN) operation is employed to explore higher-order semantic relations between different pixels within tokens to spot subtle discriminative features.  
Specifically, as shown in Fig.~\ref{fig:Flow} (A), given two adjacent tokens $\boldsymbol{T}_1$ and $\boldsymbol{T}_2$ from the transformer encoder, they are first normalized. Taking $\boldsymbol{T}_1$ as an example, it is passed through two linear projection functions (\textit{i.e.}, $\omega_v$ and $\omega_k$), respectively, to obtain the dimension-reduced feature sequences $\boldsymbol{T}_v$ and $\boldsymbol{T}_k$ ($\in \mathbb{R}^{l\times \frac{c}{2}}$), which can be denoted as $\boldsymbol{T}_v = \omega_v(\boldsymbol{T}_1)$ and $\boldsymbol{T}_k = \omega_k(\boldsymbol{T}_1)$.

Besides, $\boldsymbol{T}_1$ and $\boldsymbol{T}_2$ are concatenated to obtain an integrated token $\boldsymbol{T}_q$, which aggregates the features of both tokens, and is then exploited to interact with respective input tokens for feature enhancement. 
Specifically, another linear projection function $w_q$ is performed on this token with a dimension reduction of $c/2$, and then a softmax function is adopted to produce a weight map $T_q^w$. Next, the map is employed to weight $\boldsymbol{T}_k$ by element-wise multiplication, followed by an adaptive averaging pooling operation ($\mathcal{P}(\cdot)$) to reduce the computational costs. 
The above set of operations $\mathcal{F}_1(\cdot)$ can be denoted as: 
\begin{equation}
    T_q^{\prime} = \mathcal{F}_1(\boldsymbol{T}_k, \boldsymbol{T}_q) = \mathcal{P}( \boldsymbol{T}_k \odot \mathrm{softmax}(w_q(\boldsymbol{T}_q)) ),
\end{equation}
Then, the matrix product is applied to $\boldsymbol{T}_k$ and $\boldsymbol{T}_q^{\prime}$ to explore correlations between the two, and a softmax operation is used to generate an attention map $\boldsymbol{T}_a$, which is denoted as $\boldsymbol{T}_a = \mathrm{softmax}(\boldsymbol{T}_q^{\prime}\otimes \boldsymbol{T}_{k}^\top)$. 

After that, similar to~\cite{te2020edge}, we feed the interactive token $\boldsymbol{T}_a$ and the token $\boldsymbol{T}_v$ into the graph fusion module (GFM). In GFM, $\boldsymbol{T}_v$ is projected into the graph domain by the attention mapping $\boldsymbol{T}_a$, denoted as $\boldsymbol{T}_g = \boldsymbol{T}_v\otimes \boldsymbol{T}_a^\top$. 
In this process, a collection of pixels (``regions") with similar features are projected to one vertex, and a single-layer GCN is adopted to learn high-level semantic relations between regions and reason over non-local regions to capture global representations within tokens, by cross-vertex information propagation on the graph.
Specifically, the vertex features $\boldsymbol{T}_g$ are fed into the first-order approximation of the spectral graph convolution, and we can obtain the output $\hat{\boldsymbol{T}}_g$: 
\begin{equation}
    \hat{\boldsymbol{T}}_g = \mathrm{ReLU}((\boldsymbol{I}-\boldsymbol{A})\boldsymbol{T}_g w_g),
\end{equation}
where $\boldsymbol{A}$ is the adjacency matrix of the encoded graph connectivity and $w_g\in\mathbb{R}^{16\times 16}$ is the weight of the GCN. 

Finally, a skip connection is used to combine the input token $\boldsymbol{T}_1$ with the graph-based enhanced representation, and then a deserialization ($\mathcal{D}(\cdot)$) operation is utilized to convert the token sequences to 2D image features with the same dimension as the original features for decoding, shown as: 
\begin{equation}
     \boldsymbol{T}_{O1} = \mathcal{F}_2(\hat{\boldsymbol{T}}_g, \boldsymbol{T}_{a}, \boldsymbol{T}_1) = \mathcal{D} (\hat{\boldsymbol{T}}_g \otimes \boldsymbol{T}_{a}^\top + \boldsymbol{T}_1),
\end{equation}
where $\boldsymbol{T}_{O1}\in \mathcal{R}^{C\times \frac{H}{s}\times \frac{W}{s}}$ is the output local enhancement features from tokens. Similarly, we can also get $\boldsymbol{T}_{O2}$.

\subsection{Feature Shrinkage Decoder} 
Common decoders, as shown in Fig.~\ref{fig:DS} (a)-(d), usually directly aggregate features with significant inconsistencies, \textit{e.g.}, low-level features with rich details and high-level features with semantics, which easily introduces noise and loses subtle but valuable cues~\cite{ma2021pyramidal}. This is very unfriendly for the task of identifying camouflaged objects from inconspicuous cues. To this end, we design a feature shrinkage decoder (FSD) that progressively aggregates adjacent features in pairs using a hierarchical shrinkage pyramid architecture to accumulate more imperceptible effective cues.  
Furthermore, in our FSD decoder, we propose an adjacent interaction module (AIM) that interacts and merges the current adjacent feature pair and the aggregated features output by the previous AIM, and passes the current aggregated features to the next layer and the next AIM. It can be seen that AIM is served as a bridge for adjacent feature fusion and information passing (at the same layer and cross layer) in the decoder. 
As shown in Fig.~\ref{fig:Flow}, we can see that our decoder builds both bottom-up and left-to-right feature flows to retain more useful features. The proposed decoder can smoothly flow and accumulate the camouflaged object cues and avoid interference caused by large feature differences. 

Specifically, suppose that $\boldsymbol{F}_i$ and $\boldsymbol{F}_{i-1}$ are the adjacent feature pair of the current layer, and $\boldsymbol{F}_p$ is the output aggregated feature from the previous AIM, AIM can be formulated as: 
\begin{equation}
\begin{aligned}
    \boldsymbol{F}_{p} &= \mathrm{CBR} (\mathrm{Cat} ( \mathrm{CBR} (\mathrm{Cat} (\boldsymbol{F}_{p-1}, \boldsymbol{F}_i)), \boldsymbol{F}_{i-1}) ) \\
    \boldsymbol{F}_i^{\prime} &= \mathrm{Up} (\mathrm{CBR} (\boldsymbol{F}_{p})),
\end{aligned}
\end{equation}
where $\boldsymbol{F}_{p}$ is the feature passed to the next AIM, and $\boldsymbol{F}_i^{\prime}$ is the output feature of the current AIM for the next layer. $\mathrm{CBR}(\cdot)$ is composed of convolution, batch normalization, and ReLU operations. $\mathrm{Cat}(\cdot)$ and $\mathrm{Up}(\cdot)$ are the concatenation and $2\times$ upsampling operations, respectively. 

Note that FSD contains a total of 4 layers of shrinkage pyramid and 12 AIMs. The whole FSD process is summarized in Algorithm~\ref{alg:1}. 
The output feature from the last AIM is supervised by the ground truth ($\boldsymbol{G}$) after sigmoid and upsampling operations for camouflaged object prediction. We also supervise the output prediction ($\boldsymbol{P}_i$) at each layer of the FSD using a binary cross-entropy loss $(\mathcal{L}_{bce})$ and assign smaller weights to shallow outputs with lower detection precision. Finally, the overall loss function is: 
\begin{equation}
    \mathcal{L} _{total}=\sum_{i=0}^2{2^{(i-4)}}~\mathcal{L} _{bce}(\boldsymbol{P}_i, \boldsymbol{G}) +\mathcal{L}_{bce}(\boldsymbol{P}_3, \boldsymbol{G}),
\end{equation}
where $i$ denotes the $i$-th layer of FSD and $P_3$ means the last layer of output prediction.

\begin{algorithm}[tb]
    \algsetup{linenosize=\small} \small 
    \renewcommand{\algorithmicrequire}{\textbf{Input:}}
    \renewcommand{\algorithmicensure}{\textbf{Output:}}
    \caption{Feature shrinkage decoder.}
    \label{alg:1}
    \begin{algorithmic}[1]
	\REQUIRE $\left\{F^0_n|n\in[12,1]\right\}$, inputs for FSD layer ``0''.
	\ENSURE $\left\{P_i|i\in[0,3]\right\}$, predictions for each layer of FSD.
	\STATE The number of outputs in each layer of FSD is $num\_op=[6,3,2,1]$ 
	\STATE \textbf{for} $(i,m)$ \textbf{in} enumerate($num\_op$)~:
	\STATE ~~~~\textbf{for} $n=[m:1]$~:
	\STATE ~~~~~~~~ \textbf{if} $n=m$~:
	\STATE ~~~~~~~~~~~~~~$(dF^i_n,F^{i+1}_n)\leftarrow AIM(F^{i}_{2n},F^{i}_{2n-1})$ ;
	\STATE ~~~~~~~~ \textbf{else}~:
	\STATE ~~~~~~~~~~~~~~$(dF^i_n,F^{i+1}_n)\leftarrow AIM(\rm{Cat}(dF^i_n,F^{i}_{2n}),F^{i}_{2n-1})$ ;
	\STATE ~~~~\textbf{end for}
	\STATE ~~~~~~~~ $P_i={\rm sigmoid}(dF^i_n)$
	\STATE \textbf{end for}
    \end{algorithmic}  
\end{algorithm}

It should be noted that, unlike~\cite{ma2021pyramidal}, the proposed FSD not only adopts the cross-layer feature interaction, but also adopts the feature interaction within the same layer, to better flow and accumulate effective features in the pyramid structure, thereby minimizing the loss of subtle but crucial features in the decoder process. Furthermore, we apply lateral supervision to each layer to force each decoder layer to mine and aggregate effective camouflaged object features. Besides, to alleviate the decoder structure, the proposed decoder only integrates adjacent features without overlapping, thus reducing aggregation operations. Tab.~\ref{tab:ab2} shows the performance superiority of the proposed decoder.

\begin{table*}[]
	\centering
 	\renewcommand{\arraystretch}{1}
	\renewcommand{\tabcolsep}{1.41mm}
	\caption{Quantitative comparison with 24 SOTA methods on three benchmark datasets. Notes $\uparrow$ / $\downarrow$ denote the larger/smaller is better, respectively. ``--'' is not available. The best and second best are \textbf{bolded} and \underline{underlined} for highlighting, respectively.} \vspace{-2mm}
	\label{tab:qc}
	\scalebox{0.84}{
	\begin{tabular}{lcccccc|cccccc|cccccc}
		\toprule
		\multirow{2}{*}{\textbf{Methods}} & \multicolumn{6}{c}{\textbf{CAMO (250)}}  & \multicolumn{6}{c}{\textbf{COD10K (2,026)}}& \multicolumn{6}{c}{\textbf{NC4K (4,121)}}\\
		
		\cmidrule[0.05em](lr){2-7} \cmidrule[0.05em](lr){8-13} \cmidrule[0.05em](lr){14-19} 
		&$S_m\uparrow$&$F^{\omega}_{\beta}\uparrow$&$F^{m}_{\beta}\uparrow$&$E_{\phi}^m\uparrow$&$E_{\phi}^x\uparrow$&$\mathcal{M}\downarrow$
		&$S_m\uparrow$&$F^{\omega}_{\beta}\uparrow$&$F^{m}_{\beta}\uparrow$&$E_{\phi}^m\uparrow$&$E_{\phi}^x\uparrow$&$\mathcal{M}\downarrow$
		&$S_m\uparrow$&$F^{\omega}_{\beta}\uparrow$&$F^{m}_{\beta}\uparrow$&$E_{\phi}^m\uparrow$&$E_{\phi}^x\uparrow$&$\mathcal{M}\downarrow$ \\
		\midrule
		\multicolumn{19}{c}{Salient Object Detection} \\
		\midrule
		BASNet$_{19}$ & .618 & .413 & .475 & .661 & .708 & .159 & .634 & .365 & .417 & .678 & .735 & .105 & .695 & .546 & .610 & .762 & .786 & .095 \\
        CPD$_{19}$    & .716 & .556 & .618 & .723 & .796 & .113 & .750 & .531 & .595 & .776 & .853 & .053 & .717 & .551 & .597 & .724 & .793 & .092 \\
        EGNet$_{19}$  & .662 & .495 & .567 & .683 & .780 & .125 & .733 & .519 & .583 & .761 & .836 & .055 & .767 & .626 & .689 & .793 & .850 & .077 \\
        SCRN$_{19}$  & .779 & .643 & .705 & .797 & .850 & .090 & .789 & .575 & .651 & .817 & .880 & .047 & .830 & .698 & .757 & .854 & .897 & .059 \\
        F$^3$Net$_{20}$  & .711 & .564 & .616 & .741 & .780 & .109 & .739 & .544 & .593 & .795 & .819 & .051 & .780 & .656 & .705 & .824 & .848 & .070 \\
        CSNet$_{20}$  & .771 & .642 & .705 & .795 & .849 & .092 & .778 & .569 & .635 & .810 & .871 & .047 & .750 & .603 & .655 & .773 & .793 & .088 \\
        SSAL$_{20}$   & .644 & .493 & .579 & .721 & .780 & .126 & .668 & .454 & .527 & .768 & .789 & .066 & .699 & .561 & .644 & .780 & .812 & .093 \\
        ITSD$_{20}$   & .750 & .610 & .663 & .780 & .830 & .102 & .767 & .557 & .615 & .808 & .861 & .051 & .811 & .680 & .729 & .845 & .883 & .064 \\
        UCNet$_{20}$  & .739 & .640 & .700 & .787 & .820 & .094 & .776 & .633 & .681 & .857 & .867 & .042 & .811 & .729 & .775 & .871 & .886 & .055 \\
        VST$_{21}$    & .787 & .691 & .738 & .838 & .866 & .076 & .781 & .604 & .653 & .837 & .877 & .042 & .831 & .732 & .771 & .877 & .901 & .050 \\
		\midrule
		\multicolumn{19}{c}{Camouflaged Object Detection} \\
		\midrule
		SINet$_{20}$   & .751 & .606 & .675 & .771 & .831 & .100 & .771 & .551 & .634 & .806 & .868 & .051 & .808 & .723 & .769 & .871 & .883 & .058 \\
        SLSR$_{21}$    & .787 & .696 & .744 & .838 & .854 & .080 & .804 & .673 & .715 & .880 & .892 & .037 & .840 & .766 & .804 & .895 & .907 & .048 \\
        PFNet$_{21}$   & .782 & .695 & .746 & .842 & .855 & .085 & .800 & .660 & .701 & .877 & .890 & .040 & .829 & .745 & .784 & .888 & .898 & .053 \\
        MGL-R$_{21}$   & .775 & .673 & .726 & .812 & .842 & .088 & .814 & .666 & .711 & .852 & .890 & .035 & .833 & .740 & .782 & .867 & .893 & .052 \\
        UJSC$_{21}$    & .800 & .728 & .772 & .859 & .873 & .073 & .809 & .684 & .721 & .884 & .891 & .035 & .842 & .771 & .806 & .898 & .907 & .047 \\
        C$^2$FNet$_{21}$  & .796 & .719 & .762 & .854 & .864 & .080 & .813 & .686 & .723 & .890 & .900 & .036 & .838 & .762 & .795 & .897 & .904 & .049 \\
        UGTR$_{21}$   & .784 & .684 & .735 & .822 & .851 & .086 & .817 & .666 & .712 & .853 & .890 & .036 & .839 & .747 & .787 & .875 & .899 & .052 \\
        PreyNet$_{22}$ & .790 & .708 & .757 & .842 & .857 & .077 & .813 & .697 & .736 & .881 & .891 & .034 & -- & -- & -- & -- & -- & -- \\
        BSA-Net$_{22}$ & .794 & .717 & .763 & .851 & .867 & .079 & .818 & .699 & .738 & .891 & .901 & .034 & .841 & .771 & .808 & .897 & .907 & .048 \\
        OCE-Net$_{22}$ & .802 & .723 & .766 & .852 & .865 & .080 & .827 & .707 & .741 & .894 & .905 & .033 & \underline{.853} & .785 & .818 & .903 & .913 & .045 \\
        BGNet$_{22}$   & .812 & .749 & .789 & .870 & .882 & .073 & .831 & .722 & .753 & \textbf{.901} & \underline{.911} & .033 & .851 & \underline{.788} & \underline{.820} & \underline{.907} & \underline{.916} & .044 \\
        SegMaR$_{22}$  & .815 & \underline{.753} & \underline{.795} & .874 & .884 & .071 & .833 & .724 & .757 & \underline{.899} & .906 & .034 & .841     & .781     & .820     & .896     & .907   & .046     \\
        ZoomNet$_{22}$ & \underline{.820} & .752 & .794 & .878 & .892 & \underline{.066} & \underline{.838} & \underline{.729} & \underline{.766} & .888 & \underline{.911} & \underline{.029} & \underline{.853} & .784 & .818 & .896 & .912 & \underline{.043} \\
        SINet-v2$_{22}$ & \underline{.820} & .743 & .782 & \underline{.882} & \underline{.895} & .070 & .815 & .680 & .718 & .887 & .906 & .037 & .847 & .770 & .805 & .903 & .914 & .048 \\
        \midrule
        \textbf{Ours}    & \textbf{.856} & \textbf{.799} & \textbf{.830} & \textbf{.899} & \textbf{.928} & \textbf{.050} & \textbf{.851} & \textbf{.735} & \textbf{.769} & .895 & \textbf{.930} & \textbf{.026} & \textbf{.879} & \textbf{.816} & \textbf{.843} & \textbf{.915} & \textbf{.937} & \textbf{.035} \\
		\bottomrule
	\end{tabular}
	}
\end{table*}

\section{Experiments and Results}

\begin{figure*}[htb]
	\centering
	\begin{overpic}[width=0.94\linewidth]{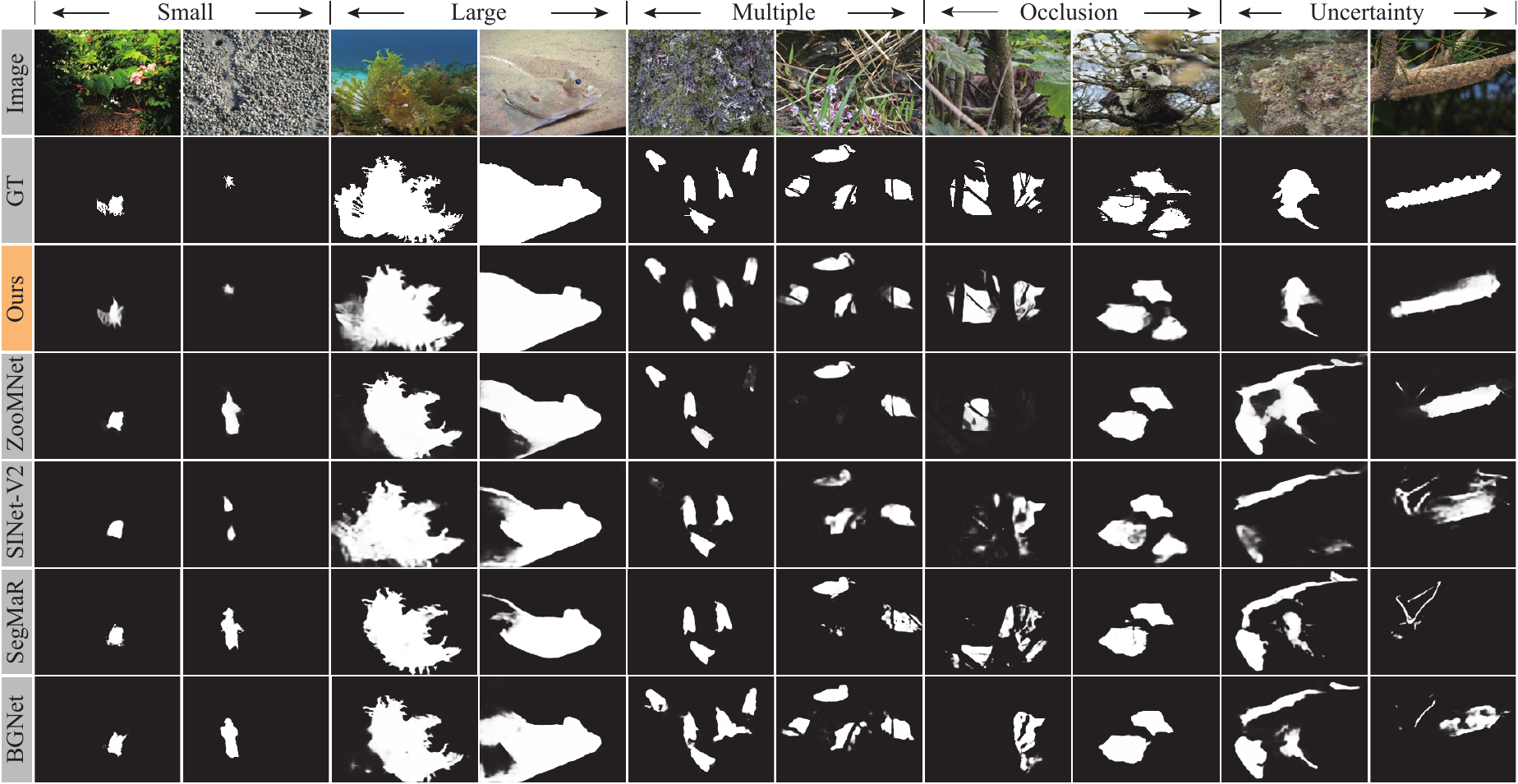}
	\end{overpic}
	\caption{\textbf{Visual comparison with some representative SOTA models in challenging scenarios.} Please zoom in for details. More visual results are provided in the \textit{Supplementary Material}.} 
	\vspace{-2mm}
	\label{fig:VC-main}
\end{figure*}

\subsection{Experiment Settings}

\noindent\textbf{Datasets.}
We evaluate the proposed method on three widely used COD datasets \textit{i.e.}, CAMO \cite{le2019anabranch},  COD10K \cite{fan2020camouflaged}, and NC4K \cite{lv2021simultaneously}. CAMO is the first COD dataset, containing 1,250 camouflaged images and 1,250 non-camouflaged images.  
COD10K is currently the largest COD dataset, which contains 5,066 camouflaged, 3,000 background, and 1,934 non-camouflaged images. NC4K is another recently released large-scale COD testing dataset that contains 4,121 images.

\noindent\textbf{Evaluation Metrics.}
We adopt six well-known evaluation metrics, including S-measure \cite{fan2017structure} ($S_m$), weighted F-measure \cite{margolin2014evaluate} ($F^{\omega}_{\beta}$), mean F-measure \cite{achanta2009frequency} ($F^{m}_{\beta}$), mean E-measure \cite{fan2018enhanced} ($E_{\phi}^m$), max E-measure ($E_{\phi}^x$), and mean absolute error ($\mathcal{M}$).

\noindent\textbf{Implementation Details.}
The proposed model is implemented by PyTorch. The base version of ViT \cite{dosovitskiy2020image}, pre-trained by the DeiT strategy \cite{touvron2021training}, is adopted as the transformer encoder. Other modules are randomly initialized. 
We follow the training set settings in \cite{youwei2022zoom,fan2021concealed} and adopt random flipping to augment the training data. All the input images are resized to 384$\times$384.
Adam is used as the optimizer, and the learning rate is initialized to 1e-4 and then scaled down by 10 every 50 epochs. 
The complete training process for 200 epochs with a batch size of 2 takes $\sim$8 hours on a workstation with 8 NVIDIA Tesla V100 GPUs.

\begin{figure}[t]
	\centering
	\begin{overpic}[width=1\linewidth]{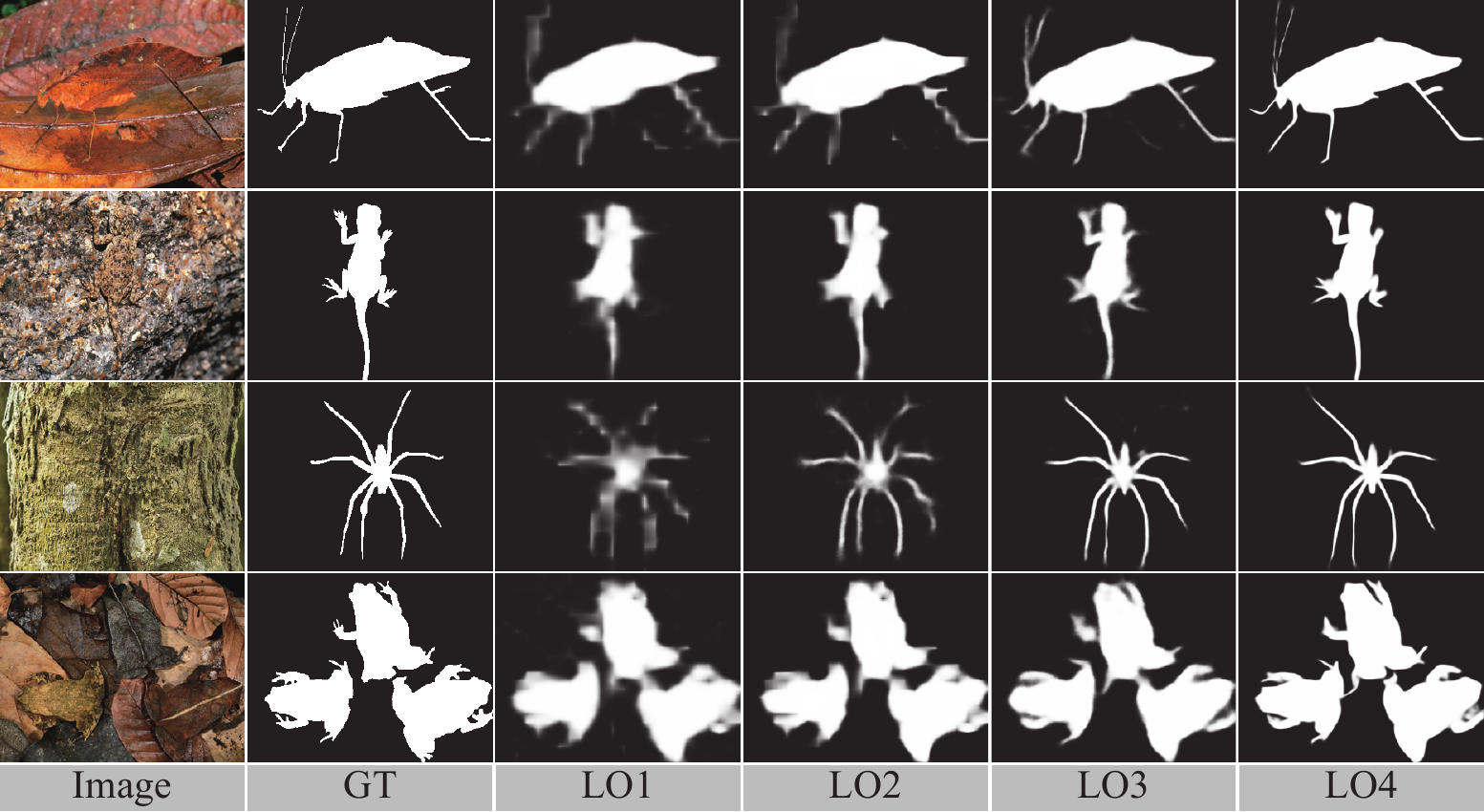}
	\end{overpic}
	\caption{\textbf{Visual comparison of the lateral output of FSD}. From LO1 to LO4 (final output) denote the layers of FSD.} 
	\label{fig:LO}
\end{figure}

\subsection{Comparison with State-of-the-Art Methods}

To demonstrate the effectiveness of the proposed method, we compare it with 24 state-of-the-art methods, including 10 salient object detection methods (\textit{i.e.}, BASNet \cite{qin2019basnet}, CPD-R \cite{wu2019cascaded}, EGNet \cite{zhao2019egnet}, SCRN \cite{wu2019stacked}, F$^3$Net \cite{wei2020f3net}, CSNet \cite{gao2020highly}, SSAL \cite{zhang2020weakly}, ITSD \cite{zhou2020interactive}, UCNet \cite{zhang2020uc}, and VST \cite{liu2021visual}), and 14 COD methods (\textit{i.e.}, SINet \cite{fan2020camouflaged}, SLSR \cite{lv2021simultaneously}, PFNet \cite{mei2021camouflaged}, MGL-R \cite{zhai2021mutual}, UJSC \cite{li2021uncertainty}, PreyNet \cite{zhang2022preynet}, BSA-Net \cite{zhu2022can}, C$^2$FNet \cite{sun2021context}, UGTR \cite{yang2021uncertainty}, OCE-Net \cite{liu2022modeling}, BGNet \cite{sun2022Boundary}, SegMaR \cite{jia2022segment}, ZoomNet \cite{youwei2022zoom}, and SINet-v2 \cite{fan2021concealed}). All the predictions of competitors are either provided by the authors or generated by models retrained based on open-source codes. More experimental results are provided in the \textit{Supplementary Material}.

\noindent\textbf{Quantitative Comparison.} 
Tab.~\ref{tab:qc} summarizes the quantitative results of our proposed method against 24 competitors on three challenging COD benchmark datasets under six evaluation metrics. It can be seen that the specially designed COD methods generally outperform the SOD models. Furthermore, our proposed method consistently surpasses all other models on these datasets. Compared to the recently proposed state-of-the-art  ZoomNet~\cite{youwei2022zoom}, our method achieves average performance gains of 3.0\%, 3.7\%, 2.7\%, 1.8\%, 3.0\%, and 17.7\% in terms of $S_\alpha$, $F^w_\beta$, $F^m_\beta$, $E_{\phi}^m$, $E_{\phi}^x$, and $\mathcal{M}$ on these three datasets, respectively. Compared to the recently proposed SINet-v2 \cite{fan2021concealed}, the average gains are 4.2\%, 7.2\%, 6.0\%, 1.4\%, 3.0\%, and 28.5\%, respectively. Besides, compared to the transformer-based methods (\textit{i.e.}, VST \cite{liu2021visual} and UGTR \cite{yang2021uncertainty}), our method shows significant performance improvements of 7.8\%, 16.3\%, 13.2\%, 6.2\%, 5.7\%, and 34.1\% over VST and 6.0\%, 12.1\%, 9.3\%, 6.3\%, 5.9\%, and 34.1\% over UTGR on average for $S_\alpha$, $F^w_\beta$, $F^m_\beta$, $E_{\phi}^m$, $E_{\phi}^x$, and $\mathcal{M}$, respectively. 
The superiority in performance benefits from the compensation of the local feature modeling for the transformer backbones, and the smooth and progressive feature decoding to accumulate more subtle clues of the camouflage objects.

\begin{table*}[tb]
	\centering
	\renewcommand{\tabcolsep}{1.32mm}
	\caption{Ablation studies of FSPNet on benchmark datasets. ``B" is backbone, ``D" is FSD and ``T" is NL-TEM.} \vspace{-2mm}
	\label{tab:ab1}
	\scalebox{0.84}{
	\begin{tabular}{lcccccc|cccccc|cccccc}
		\toprule
		\multirow{2}{*}{\textbf{Settings}} & \multicolumn{6}{c}{\textbf{CAMO (250)}}  & \multicolumn{6}{c}{\textbf{COD10K (2,026)}}& \multicolumn{6}{c}{\textbf{NC4K (4,121)}}\\
		
		\cmidrule[0.05em](lr){2-7} \cmidrule[0.05em](lr){8-13} \cmidrule[0.05em](lr){14-19} 
		&$S_m\uparrow$&$F^{\omega}_{\beta}\uparrow$&$F^{m}_{\beta}\uparrow$&$E_{\phi}^m\uparrow$&$E_{\phi}^x\uparrow$&$\mathcal{M}\downarrow$
		&$S_m\uparrow$&$F^{\omega}_{\beta}\uparrow$&$F^{m}_{\beta}\uparrow$&$E_{\phi}^m\uparrow$&$E_{\phi}^x\uparrow$&$\mathcal{M}\downarrow$
		&$S_m\uparrow$&$F^{\omega}_{\beta}\uparrow$&$F^{m}_{\beta}\uparrow$&$E_{\phi}^m\uparrow$&$E_{\phi}^x\uparrow$&$\mathcal{M}\downarrow$ \\ 
		\midrule
        $\rm{B_1} $  & .774 & .685 & .732 & .813 & .839 & .089 & .791 & .659 & .709 & .855 & .882 & .042 & .828 & .747 & .795 & .881 & .901 & .051 \\
        $\rm{B_4} $      & .781 & .693 & .738 & .818 & .845 & .086 & .801 & .668 & .713 & .861 & .889 & .040 & .835 & .755 & .802 & .885 & .909 & .049 \\
        $\rm{B_8} $      & .795 & .715 & .746 & .827 & .856 & .081 & .807 & .679 & .725 & .867 & .897 & .039 & .841 & .766 & .807 & .887 & .913 & .048 \\
        $\rm{B_{12}} $    & .798 & .726 & .755 & .837 & .868 & .079 & .812 & .697 & .732 & .871 & .901 & .038 & .853 & .781 & .813 & .901 & .918 & .046 \\
        \midrule
        $\rm{B_{4}+D}$   & .786 & .716 & .743 & .831 & .857 & .082 & .830 & .696 & .735 & .873 & .894 & .038 & .854 & .773 & .810 & .902 & .921 & .048 \\
        $\rm{B_{8}+D}$    & .807 & .731 & .759 & .839 & .871 & .076 & .831 & .711 & .740 & .882 & .912 & .036 & .866 & .787 & .823 & .905 & .922 & .043 \\
        $\rm{B_{12}+D}$   & .817 & .755 & .786 & .858 & .891 & .062 & .844 & .728 & .759 & .888 & .918 & .033 & .870 & .808 & .836 & .912 & .929 & .040 \\
        \midrule
        $\rm{B_{4}+D+T}$ & .809 & .731 & .762 & .837 & .877 & .078 & .836 & .707 & .742 & .883 & .913 & .036 & .864 & .786 & .824 & .906 & .927 & .038 \\
        $\rm{B_{8}+D+T}$  & .827 & .762 & .788 & .862 & .912 & .066 & .842 & .724 & .757 & .887 & .922 & .029 & .868 & .803 & .832 & .907 & .932 & .037 \\
        $\bm{{\rm B_{12}+D+T}}$ & \textbf{.856} & \textbf{.799} & \textbf{.830} & \textbf{.899} & \textbf{.928 }& \textbf{.050} & \textbf{.851} & \textbf{.735} & \textbf{.769} & \textbf{.895} & \textbf{.930} & \textbf{.026} & \textbf{.879} & \textbf{.816} & \textbf{.843} & \textbf{.915} & \textbf{.937} & \textbf{.035} \\
		\bottomrule
	\end{tabular}
	}
\end{table*}

\begin{table}[tb]
	\centering
	\renewcommand{\tabcolsep}{1.0mm}
	\caption{More ablation studies on COD10K and NC4K.} \vspace{-2mm}
	\label{tab:ab2}
	\scalebox{0.65}{
    \begin{tabular}{ccccccc|cccccc}
    \toprule
    \multirow{2}{*}{\textbf{No.}} & \multicolumn{6}{c}{\textbf{COD10K}} & \multicolumn{6}{c}{\textbf{NC4K}} \\
    \cmidrule[0.05em](lr){2-7} \cmidrule[0.05em](lr){8-13}
            &$S_m\uparrow$&$F^{\omega}_{\beta}\uparrow$&$F^{m}_{\beta}\uparrow$&$E_{\phi}^m\uparrow$&$E_{\phi}^x\uparrow$&$\mathcal{M}\downarrow$& $S_m\uparrow$&$F^{\omega}_{\beta}\uparrow$&$F^{m}_{\beta}\uparrow$&$E_{\phi}^m\uparrow$&$E_{\phi}^x\uparrow$&$\mathcal{M}\downarrow$ \\
    \midrule
    {\Large\ding{172}}   & .825  & .710   & .739 & .875 & .908 & .037   & .859  & .786   & .820 & .904 & .920 & .044   \\
    {\Large\ding{173}}   & .848  & .731   & .764 & .891 & .923 & .027   & .875  & .811   & .837 & .910 & .924 & .037   \\
    {\Large\ding{174}}   & .840  & .722   & .753 & .882 & .916 & .034   & .867  & .798   & .832 & .906 & .921 & .039   \\
    {\Large\ding{175}}  & .849 &  .732 &  .761  & .887 & .922 & .029 & .872   & .804  &  .832  & .901 &  .927 & .038  \\
    \textbf{Ours}  & .851  & .735   & .769 & .895 & .930 & .026   &  .879 & .816   & .843 & .915 & .937 &.035    \\
    \midrule
    {\Large\ding{176}}  & .844  &  .728  & .759 & .888 & .918 & .033   & .870  & .808 & .836 & .912 & .929 & .040\\
    +GFM   & .846  & .732   & .764 & .889 & .924 & .028   & .874  & .810   & .838 & .913 & .925 & .037   \\
    +NL   & .847  & .731   & .765 & .892 & .926 & .028   & .873  & .811   & .839 & .913 & .926 & .036   \\
    \bottomrule
    \end{tabular}
}
\end{table}

\noindent\textbf{Visual Comparison.} 
Fig. \ref{fig:VC-main} shows the visual comparisons of our proposed method with some representative competitors in several typical scenarios, including small, large, multiple, occluded objects, and uncertain boundaries. It can be seen that the compared methods are prone to provide inaccurate object localization, incomplete object regions, or missing objects, resulting in inferior segmentation of camouflaged objects. Our proposed method shows superior visual performance for more accurate and complete predictions. Experiments also demonstrate the robustness of the proposed method to different challenging scenarios.

\subsection{Ablation Study}
To validate the effectiveness of the proposed modules for COD, we perform the following ablation studies on these COD benchmark datasets.

\noindent\textbf{Dense Integration Strategy.}
Integrating multiple backbone features to improve prediction is widely used in segmentation tasks. Therefore, we test combinations of different lateral features of the base ViT (denoted as B) for decoding. The baseline decoder contains concatenation, reshape, and upsampling operations.
The results are shown in the first four rows of Tab.~\ref{tab:ab1}, where $i$ in B$_i$ denotes the number of feature layers adopted for decoding. We can see that aggregating different feature layers benefit merging more clues, thereby improving the detection performance. In our experiments, aggregating all the transformer feature layers (\textit{i.e.}, B$_{12}$) provided the best performance.

\noindent\textbf{Feature Shrinkage Decoder.}
Tab.~\ref{tab:ab1} (5th$\sim$7th) shows the results of our proposed decoder FSD (denoted as D) under different backbone feature combinations. Note that the number of pyramid layers is 2, 3, and 4 for ``(B$_4$+D)", ``(B$_8$+D)", and ``(B$_{12}$+D)", respectively. 
We can see that FSD effectively improves the performance, showing  the designed FSD well aggregates and retains critical features of different layers for accurate predictions. 
Moreover, Fig.~\ref{fig:LO} provides the outputs of different pyramidal decoder layers in FSD (from LO1 to LO4 by depth), validating the ability of FSD to recover object details and generate clear predictions gradually. 

Besides, we conducted five experiments to verify the effectiveness of the decoder components and structures, including {\ding{172}} replacing FSD with a U-shaped decoding structure (similar to Fig.~\ref{fig:DS} (a)), {\ding{173}} replacing AIM with a simpler combination of operations (\textit{i.e.}, concatenation and $1\times1$ convolution), {\ding{174}} extending AIM to aggregate three adjacent feature layers, {\ding{175}} adjusting our decoder to pairwise feature aggregation with overlap and removing lateral supervision and feature interaction within the same layer (similar to \cite{ma2021pyramidal}).  
Note that we retain other modules in experiments. The results are shown in Tab.~\ref{tab:ab2} (1st$\sim$5th rows).

Our decoder and {\ding{175}} outperforms the U-shaped decoding structure ({\ding{172}}) by a large margin, this is because this type of decoder usually directly aggregates features (in the same fusion layer) with large feature differences, and tends to discard some subtle but valuable cues, resulting in inaccurate predictions, especially for the task of identifying camouflaged objects from faint clues. Our decoder and {\ding{175}} both progressively aggregate adjacent features through a layer-by-layer shrinkage pyramid (multiple fusion layers) to accumulate valuable cues as much as possible for object prediction. However, our decoder introduces lateral supervision and feature flow within the same layer, which forces the decoder to accumulate more critical camouflaged object cues, thus achieving a large performance improvement, especially on the NC4K dataset, compared to {\ding{175}}. Besides, by comparing with {\ding{173}} and {\ding{174}}, the proposed AIM provides better performance for camouflaged object prediction.

\noindent\textbf{Non-local Token Enhancement Module.} Tab.~\ref{tab:ab1} (8th$\sim$10th rows) shows the results of NL-TEM (denoted as $T$). The NL-TEM complements the local feature exploration for transformers, which contributes to the recovery of objects' local details, and further improves the performance. 

Besides, we perform two additional experiments to verify the effectiveness of non-local operations and graph convolutions. The results are shown in Tab.~\ref{tab:ab2} (6th$\sim$8th rows). 
{\ding{176}} denotes ``B$_{12}$+D". Based on model {\ding{176}}, we add the GFM module (``+GFM"), that is,  the two inputs of NL-TEM are directly fed into the GFM after reshape, concatenation and softmax operations. The ``+NL" denotes removing the GFM directly from NL-TEM to test the non-local operations because the size of the input and output of the GFM are the same. 
It can be seen that the addition of non-local operations and GFM both contribute to camouflaged object detection and promote the improvement of detection performance. When combining these two components (\textit{i.e.}, ``B$_{12}$+D+T" in Tab.~\ref{tab:ab1}), the proposed model significantly improves the performance for camouflaged object detection.

\section{Conclusion}
Considering the existing COD methods suffer from two issues, that is, less effective locality modeling for transformer-based models and limitations of feature aggregation in decoders, in this paper, we propose a novel transformer-based feature shrinkage pyramid network (FSPNet), which contains a non-local token enhancement module (NL-TEM) and a feature shrinkage decoder (FSD) with adjacent interaction modules (AIM). The proposed model can hierarchically aggregate locality-enhanced neighboring features through progressive shrinking, thereby integrating subtle but effective local and global cues as much as possible for accurate and complete camouflaged object detection. Extensive comparison experiments and ablation studies show that the proposed FSPNet achieves superior performance over 24 cutting-edge approaches on three widely-used COD benchmark datasets. 

\vspace{2pt}
\noindent\textbf{Acknowledgments.} We thank Deng-Ping Fan for insightful feedback.


{\small
\bibliographystyle{ieee_fullname}
\balance
\bibliography{egbib}
}


\newpage

\section*{\Large Appendices}
We will introduce more details that cannot be expanded in the main text.

\section{Datasets and Evaluation Details}
\subsection{Datasets}
We conduct experiments on three COD datasets, including CAMO \cite{le2019anabranch}, COD10K \cite{fan2020camouflaged}, and NC4K \cite{lv2021simultaneously}. 

\begin{itemize}
    \item \textbf{CAMO} is the first dataset for COD and contains 2.5K images (2K for training and 0.5K for testing) with manual annotations, of which 1.25K camouflaged images and 1.25K non-camouflaged images.

    \item \textbf{COD10K} is currently the largest challenging dataset for COD. It contains 10K images with dense annotations (6K for training and 4K for testing) covering 78 categories.

    \item \textbf{NC4K} is currently the largest test dataset for COD, containing 4,121 images. It not only contains binary ground truth maps, but also provides camouflaged object ranking annotations.
\end{itemize}

\subsection{Evaluation Metrics}
In our experiments, we adopt five kinds of evaluation metrics that are widely used in COD tasks for quantitative evaluation, including S-measure \cite{fan2017structure} ($S_m$), F-measure \cite{achanta2009frequency} ($F_{\beta}$), weighted F-measure \cite{margolin2014evaluate} ($F^{\omega}_{\beta}$), E-measure \cite{fan2018enhanced} ($E_m$), and mean absolute error ($MAE$, $\mathcal{M}$).

\textbf{{Mean absolute error.}} $\mathcal{M} $ is to calculate the average absolute error of the prediction of camouflaged objects ($C$) and ground truth ($G$), which is defined as:
\begin{equation}
\mathcal{M}=\frac{1}{N}\sum_{i=1}^{N}\left|C\left(i\right)-G\left(i\right) \right|,
\end{equation}
where $ N $ is the total pixels of the image.

\textbf{{S-measure.}} Considering that camouflaged objects have complex shapes, we use $ S_m $ that combines region-aware ($ S_r $) and object-aware ($S_o$) to calculate structural similarity, which is defined as:
\begin{equation}
 S_{m}=\alpha * S_{o}+\left(1-\alpha\right)*S_{r},
\end{equation}
where $\alpha \in [0,1]$ is the balance parameter and is set to 0.5 in our experiments. 

\textbf{{F-measure.}} $F_\beta$ is a comprehensive metric that takes into account both precision $ \left( P\right)  $ and recall $ \left( R\right)  $, and is defined as:
 \begin{equation}
 	F_{\beta}=\frac{\left(1+\beta ^2\right)P\cdot R}{\beta ^{2}\cdot P+R},
 \end{equation}
 where $\beta$ is the balance parameter and $\beta^2$ is set to 0.3 in this paper. We report adaptive F-measure ($F^{a}_{\beta}$), mean F-measure ($F^{m}_{\beta}$) and maximum F-measure ($F^{x}_{\beta}$) in our experiments. The ``adaptive'' means that two times the average value of the prediction map pixels is used as the threshold for calculating precision and recall. 
 
\textbf{{Weighted F-measure.}} $F^{\omega}_{\beta}$ is obtained based on the $F_{\beta}$ by combining the weighted precision defined by measure exactness and the weighted recall defined by measure completeness, which is calculated as:
\begin{equation}\label{equ:2-21}
	F_{\beta} ^{\,\omega}=\frac{\left( 1+\beta ^2 \right) \times P ^{\,\omega}\times R ^{\,\omega}}{\beta ^2\times P ^{\,\omega}+\mathrm{R}^{\mathrm{\omega}}},
\end{equation} 
 
\textbf{{E-measure.}} $ E_m  $ is a metric based on human visual perception, which can complete pixel-level matching and image-level statistics, which is denoted as:
\begin{equation}
E_{m}=\frac{1}{N}\sum_{i=1}^{N}\phi_{FM}\left(i\right),
\end{equation}
where $ \phi_{FM} $ denotes the enhanced-alignment matrix. We report adaptive E-measure ($E^{a}_{\beta}$), mean E-measure ($E^{m}_{\beta}$) and maximum E-measure ($E^{x}_{\beta}$) in our experiments.

\section{More Comparisons}

\subsection{Quantitative Experiments} 
We show more quantitative experimental results on three benchmark COD datasets. The methods used in the experiments for comparison include 10 SOD methods (BASNet  \cite{qin2019basnet}, CPD \cite{wu2019cascaded}, EGNet \cite{zhao2019egnet}, SCRN \cite{wu2019stacked}, F$^3$Net \cite{wei2020f3net}, CSNet \cite{gao2020highly}, 
SSAL \cite{zhang2020weakly}, ITSD \cite{zhou2020interactive}, UCNet \cite{zhang2020uc}, VST~\cite{liu2021visual}) and 13 COD methods (SINet \cite{fan2020camouflaged}, SLSR \cite{lv2021simultaneously}, PFNet \cite{mei2021camouflaged}, MGL-R \cite{zhai2021mutual}, UJSC \cite{li2021uncertainty}, C$^2$FNet~\cite{sun2021context}, UGTR~\cite{yang2021uncertainty}, BSA-Net~\cite{zhu2022can}, OCE-Net \cite{liu2022modeling}, BGNet~\cite{sun2022Boundary}, SegMaR~\cite{jia2022segment}, ZoomNet \cite{youwei2022zoom}, SINet-v2 \cite{fan2021concealed}).

\vspace{4pt}
\noindent\textbf{Comprehensive Evaluation.}
As shown in Tab.~\ref{tab:qc1} and Tab.~\ref{tab:qc2}, we further list more comprehensive evaluation results on three COD datasets. It can be seen that our model achieves the best detection performance overall.

\vspace{4pt}
\noindent\textbf{Evaluation for Subclasses.}
In addition to the overall quantitative comparison of the COD10K dataset, we also report quantitative results of some representative competitors on each subclass in Tab. \ref{tab:subclass}. 
It can be seen that our model outperforms other competitors for most subclasses of the COD10K dataset. On the other hand, adapting and improving the model based on the results of each subclass is one of our future work.

\subsection{Qualitative Comparison}
\begin{figure*}[thp!]
	\centering
	\begin{overpic}[width=0.88\linewidth]{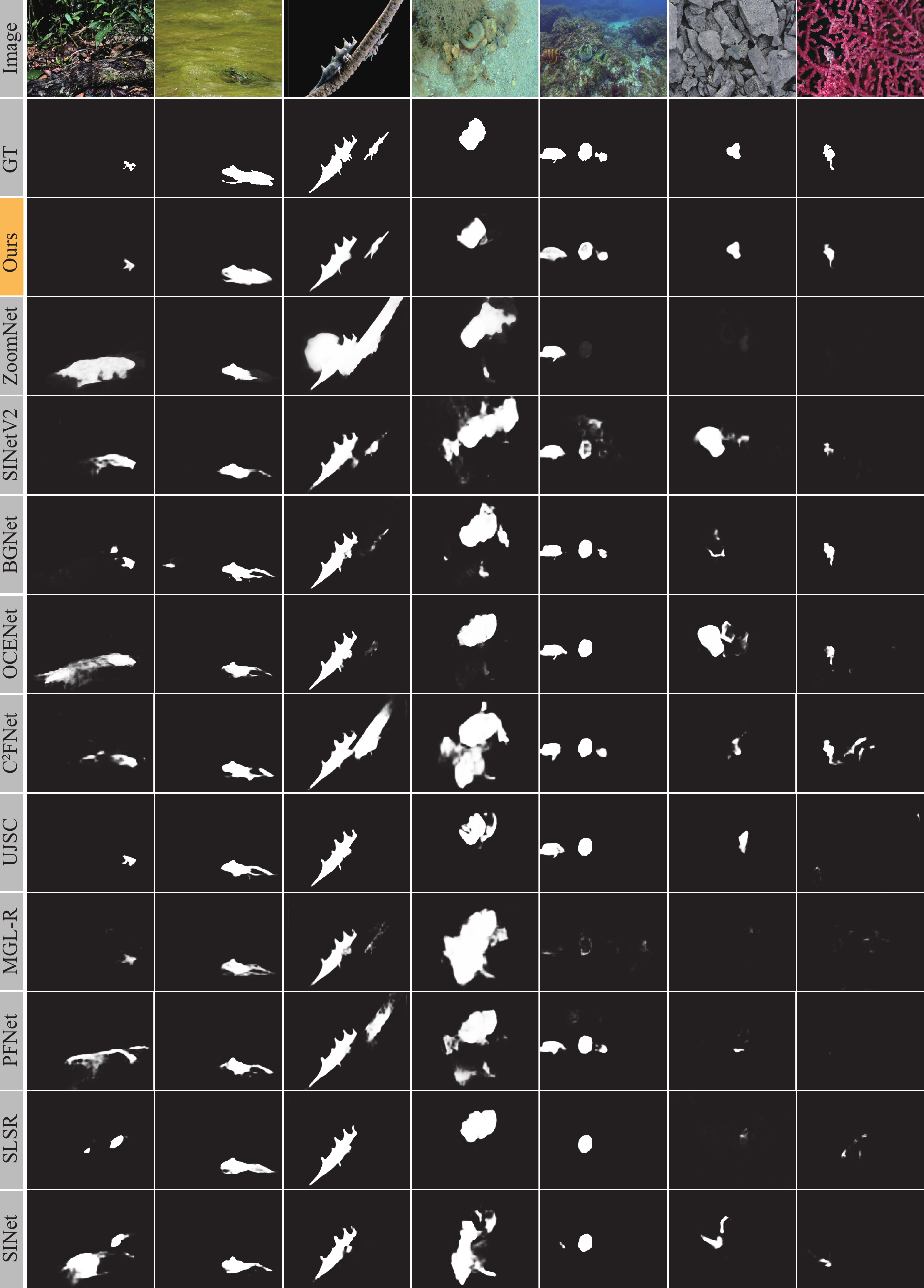}
	\end{overpic}
	\caption{Visual comparison with other competitors in detecting \textbf{small} camouflaged objects. Please zoom in for details.
	}
	\label{fig:VC-small}
\end{figure*}

\begin{figure*}[thp!]
	\centering
	\begin{overpic}[width=0.88\linewidth]{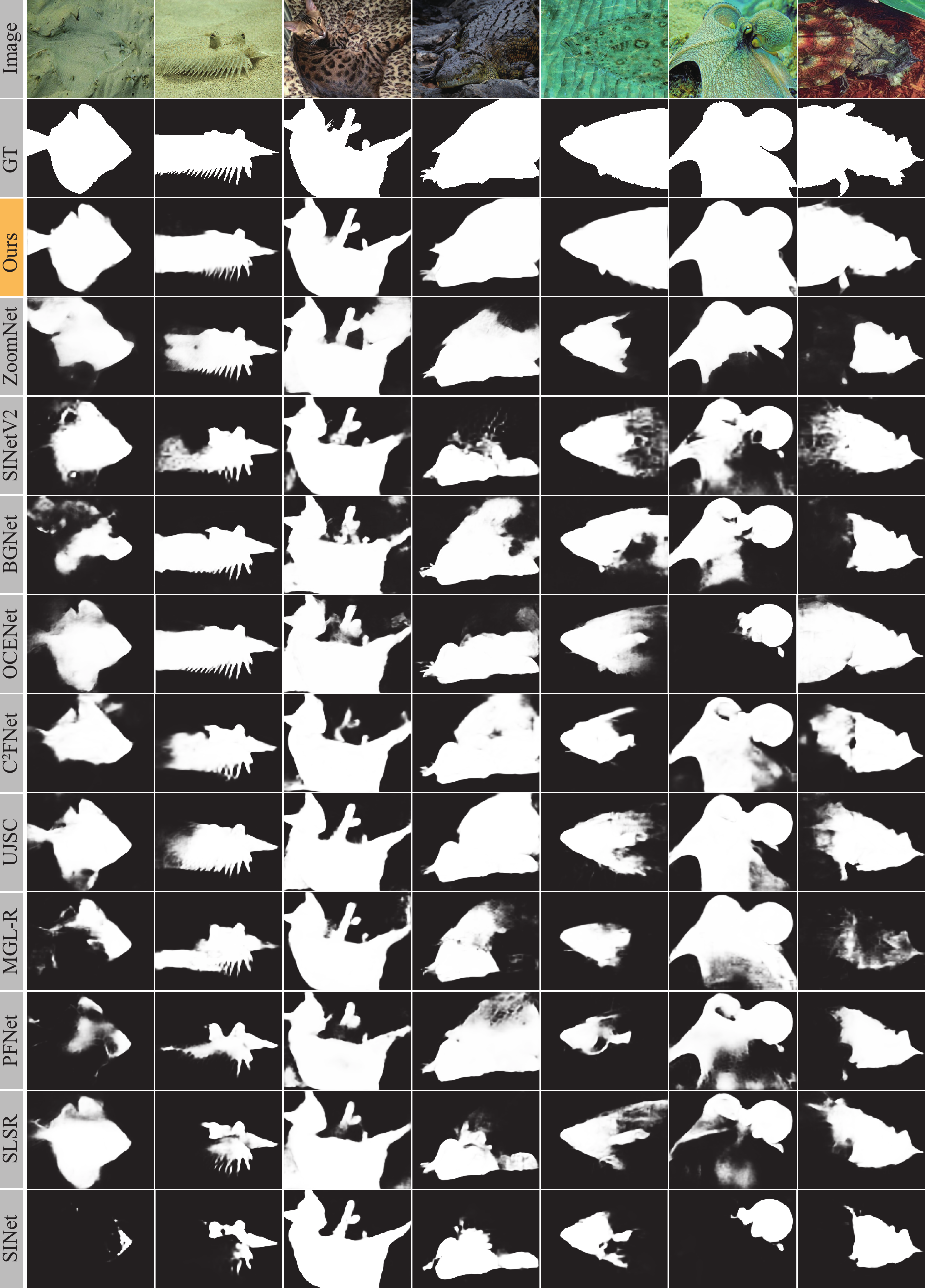}
	\end{overpic}
	\caption{Visual comparison with other competitors in detecting \textbf{big} camouflaged objects. Please zoom in for details.
	}
	\label{fig:VC-big}
\end{figure*}

\begin{figure*}[thp!]
	\centering
	\begin{overpic}[width=0.88\linewidth]{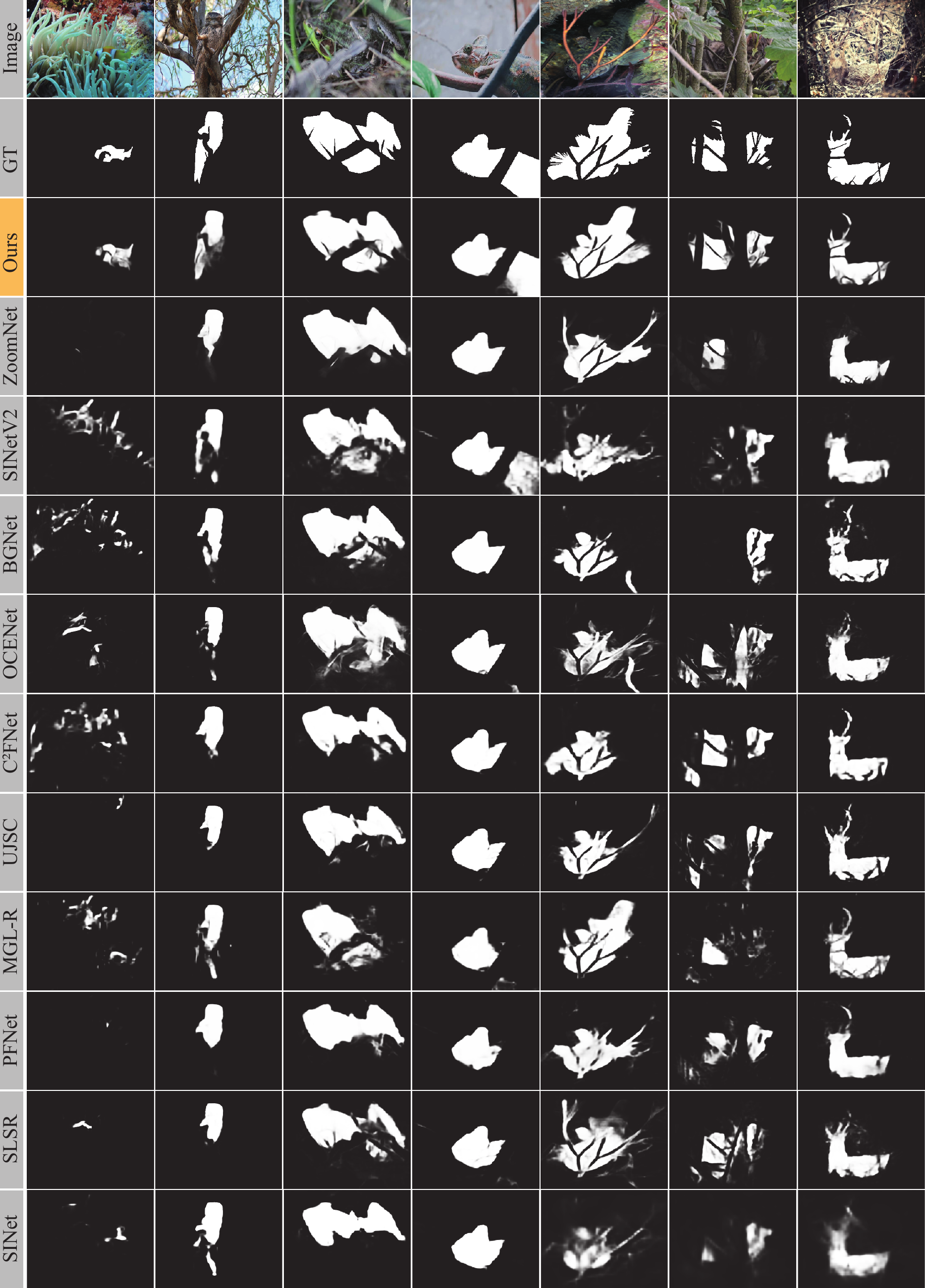}
	\end{overpic}
	\caption{Visual comparison with other competitors in detecting \textbf{obscured} camouflaged objects. Please zoom in for details.
	}
	\label{fig:VC-cover}
\end{figure*}

\begin{figure*}[thp!]
	\centering
	\begin{overpic}[width=0.88\linewidth]{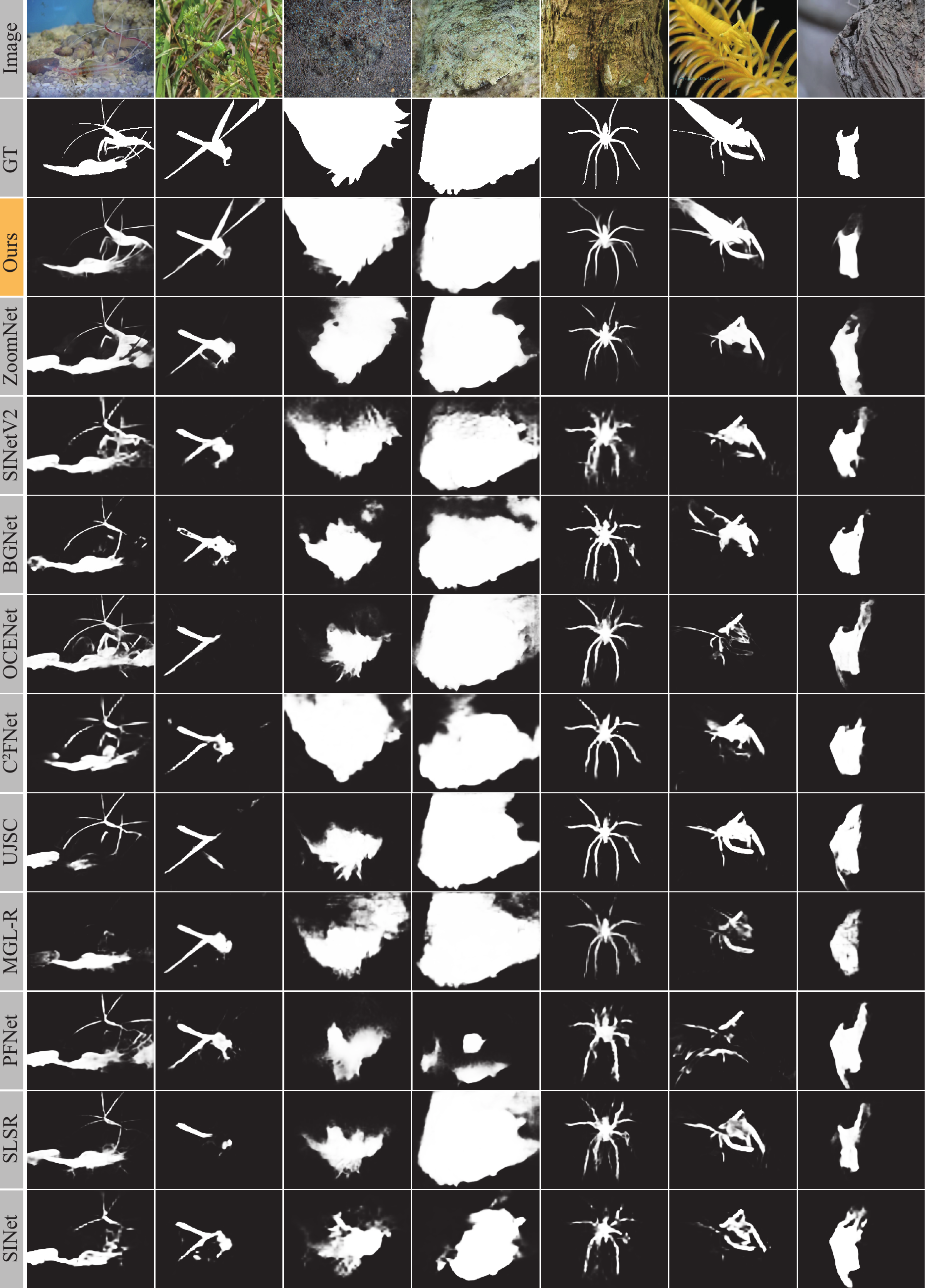}
	\end{overpic}
	\caption{Visual comparison with other competitors in detecting camouflaged objects with \textbf{indistinguishable boundaries}. Please zoom in for details.
	}
	\label{fig:VC-blur}
\end{figure*}

Due to the space limitations of the manuscript, we add more visual comparisons to this supplementary material for further demonstration of the performance of our model. Fig \ref{fig:VC-small}, \ref{fig:VC-big}, \ref{fig:VC-cover}, and \ref{fig:VC-blur} show examples containing small, large, obscured, and boundary indistinguishable camouflaged objects, respectively. As can be seen from these visual comparisons, our model is more robust to a wide range of challenging scenarios, showing superior visual performance for more accurate and complete predictions.

\begin{table}[tb]
	\centering
	\renewcommand{\arraystretch}{1.3}
	\renewcommand{\tabcolsep}{1.0mm}
	\caption{Ablation studies on COD10K and NC4K. {\ding{172}} is similar to \cite{ma2021pyramidal}, which adjusts our decoder to pairwise feature aggregation with overlap, and removes lateral supervision and feature interaction within the same layer. {\ding{173}} is a decoder that adds the lateral supervision and feature interaction within the same layer to {\ding{172}}. The difference between ours and {\ding{173}} is that our method is pairwise feature aggregation without overlapping.}
	\label{tab:ablation_FSD}
	\scalebox{0.65}{
    \begin{tabular}{ccccccc|cccccc}
    \toprule
    \multirow{2}{*}{\textbf{No.}} & \multicolumn{6}{c|}{\textbf{COD10K}} & \multicolumn{6}{c}{\textbf{NC4K}} \\
    \cmidrule[0.05em](lr){2-7} \cmidrule[0.05em](lr){8-13}
            &$S_m\uparrow$&$F^{\omega}_{\beta}\uparrow$&$F^{m}_{\beta}\uparrow$&$E_{\phi}^m\uparrow$&$E_{\phi}^x\uparrow$&$\mathcal{M}\downarrow$& $S_m\uparrow$&$F^{\omega}_{\beta}\uparrow$&$F^{m}_{\beta}\uparrow$&$E_{\phi}^m\uparrow$&$E_{\phi}^x\uparrow$&$\mathcal{M}\downarrow$ \\
    \midrule
    {\Large\ding{172}}  & .849 &  .732 &  .761  & .887 & .922 & .029 & .872   & .804  &  .832  & .901 &  .927 & .038  \\
    {\Large\ding{173}}    &  .850 &  \textbf{.736}  & .768 & .893 & \textbf{.931} & \textbf{.026}   & .878  &  \textbf{.817}  & .841 &  .912 & .936 & .036 \\
    \textbf{Ours}  & \textbf{.851}  & .735   & \textbf{.769} & \textbf{.895} & .930 & \textbf{.026}   &  \textbf{.879} & .816   & \textbf{.843} & \textbf{.915} & \textbf{.937} & \textbf{.035}    \\
    \bottomrule
    \end{tabular}
}
\end{table}

\begin{table*}[]
	\centering
 	\renewcommand{\arraystretch}{1.2}
	\renewcommand{\tabcolsep}{1.6mm}
	\caption{Quantitative comparison with 23 SOTA methods on CAMO \cite{le2019anabranch} dataset. Notes $\uparrow$ / $\downarrow$ denote the larger/smaller is better, respectively. The best and second best are \textbf{bolded} and \underline{underlined} for highlighting, respectively.} \vspace{-2mm}
	\label{tab:qc1}
	\scalebox{0.8}{
	\begin{tabular}{lccccccccc}
		\toprule
		\multirow{2}{*}{\textbf{Methods}} & \multicolumn{9}{c}{\textbf{CAMO (250)}}  \\
		
		\cmidrule[0.05em](lr){2-10} 
		&$S_m\uparrow$
		&$F^{\omega}_{\beta}\uparrow$
		&$E^{a}_m\uparrow$
		&$E^{m}_m\uparrow$
		&$E^{x}_m\uparrow$
		&$F^{a}_{\beta}\uparrow$
		&$F^{m}_{\beta}\uparrow$
		&$F^{x}_{\beta}\uparrow$
		&$\mathcal{M}\downarrow$

		\\

		\midrule
		\multicolumn{10}{c}{\textbf{Salient Object Detection}} \\
		\midrule
		BASNet$_{19}$  & .618 & .413 & .719 & .661 & .708 & .525 & .475 & .519 & .159   \\
        CPD$_{19}$  & .716 & .556 & .807 & .723 & .796 & .675 & .618 & .658 & .113  \\
        EGNet$_{19}$   & .662 & .495 & .780 & .683 & .780 & .640 & .567 & .625 & .125  \\
        SCRN$_{19}$ & .779 & .643 & .848 & .797 & .850 & .733 & .705 & .738 & .090  \\
        F$^3$Net$_{20}$   & .711 & .564 & .802 & .741 & .780 & .661 & .616 & .630 & .109  \\
        CSNet$_{20}$   & .771 & .642 & .847 & .795 & .849 & .730 & .705 & .740 & .092  \\
        SSAL$_{20}$ & .644 & .493 & .765 & .721 & .780 & .605 & .579 & .601 & .126  \\
        ITSD$_{20}$ & .750 & .610 & .830 & .780 & .830 & .692 & .663 & .694 & .102  \\
        UCNet$_{20}$   & .739 & .640 & .811 & .787 & .820 & .716 & .700 & .708 & .094  \\
        VST$_{21}$   & .787 & .691 & .866 & .838 & .866 & .746 & .738 & .756 & .076  \\
		\midrule
   \multicolumn{10}{c}{\textbf{Camouflaged Object Detection}} \\
   \midrule
        SINet$_{20}$    & .751 & .606 & .834 & .771 & .831 & .709 & .675 & .706 & .100  \\
        SLSR$_{21}$     & .787 & .696 & .855 & .838 & .854 & .756 & .744 & .753 & .080  \\
        PFNet$_{21}$    & .782 & .695 & .852 & .842 & .855 & .751 & .746 & .758 & .085  \\
        MGL-R$_{21}$    & .775 & .673 & .847 & .812 & .842 & .738 & .726 & .740 & .088  \\
        UJSC$_{21}$     & .800 & .728 & .865 & .859 & .873 & .779 & .772 & .779 & .073  \\
        C$^2$FNet$_{21}$   & .796 & .719 & .864 & .854 & .864 & .764 & .762 & .771  & .088\\
        UGTR$_{21}$     & .784 & .684 & .856 & .822 & .851 & .748 & .735 & .751 & .086  \\
        BSA-Net$_{22}$  & .794 & .717 & .859 & .851 & .867 & .768 & .763 & .770 & .079   \\
        OCE-Net$_{22}$  & .802 & .723 & .863 & .852 & .865 & .776 & .766 & .777 & .080  \\
        BGNet$_{22}$    & .812 & .749 & .876 & .870 & .882 & .786 & .789 & .799 & .073  \\
        SegMaR$_{22}$   & .815 & \underline{.753} & .872 & .874 & .884 & \underline{.795} & \underline{.795} & .803 & .071  \\
        ZoomNet$_{22}$  & \underline{.820} & .752 & \underline{.878} & .878 & .892 & .792 & .794 & \underline{.805} & \underline{.066}  \\
        SINet-v2$_{22}$ & \underline{.820} & .743 & .875 & \underline{.882} & \underline{.895} & .779 & .782 & .801 & .070  \\
   \midrule
   Ours & \textbf{.856} & \textbf{.799} & \textbf{.919} & \textbf{.899} & \textbf{.928} & \textbf{.829} & \textbf{.830} & \textbf{.846} & \textbf{.050} \\
   \bottomrule
	\end{tabular}
	}
\end{table*}

\begin{table*}[]
	\centering
 	\renewcommand{\arraystretch}{1.2}
	\renewcommand{\tabcolsep}{1.2mm}
	\caption{Quantitative comparison with 23 SOTA methods on COD10K \cite{fan2020camouflaged} and NC4K \cite{lv2021simultaneously} datasets. Notes $\uparrow$ / $\downarrow$ denote the larger/smaller is better, respectively. ``--'' is not available. The best and second best are \textbf{bolded} and \underline{underlined} for highlighting, respectively.} \vspace{-2mm}
	\label{tab:qc2}
	\scalebox{0.75}{
	\begin{tabular}{lccccccccc|ccccccccc}
		\toprule
		\multirow{2}{*}{\textbf{Methods}} & \multicolumn{9}{c}{\textbf{COD10K (2,026)}}  & \multicolumn{9}{c}{\textbf{NC4K (4,121)}} \\
		
		\cmidrule[0.05em](lr){2-10} \cmidrule[0.05em](lr){11-19}
		&$S_m\uparrow$
		&$F^{\omega}_{\beta}\uparrow$
		&$E^{a}_m\uparrow$
		&$E^{m}_m\uparrow$
		&$E^{x}_m\uparrow$
		&$F^{a}_{\beta}\uparrow$
		&$F^{m}_{\beta}\uparrow$
		&$F^{x}_{\beta}\uparrow$
		&$\mathcal{M}\downarrow$
		&$S_m\uparrow$
		&$F^{\omega}_{\beta}\uparrow$
		&$E^{a}_m\uparrow$
		&$E^{m}_m\uparrow$
		&$E^{x}_m\uparrow$
		&$F^{a}_{\beta}\uparrow$
		&$F^{m}_{\beta}\uparrow$
		&$F^{x}_{\beta}\uparrow$
		&$\mathcal{M}\downarrow$
		\\

		\midrule
		\multicolumn{19}{c}{\textbf{Salient Object Detection}} \\
		\midrule
		BASNet$_{19}$  & .634 & .365 & .676 & .678 & .735 & .421 & .417 & .451 & .105 & .695 & .546 & .784 & .762 & .786 & .618 & .610 & .617 & .095 \\
       CPD$_{19}$ & .750 & .531 & .792 & .776 & .853 & .578 & .595 & .640 & .053 & .717 & .551 & .808 & .724 & .793 & .660 & .597 & .638 & .092 \\
       EGNet$_{19}$ & .733 & .519 & .799 & .761 & .836 & .572 & .583 & .620 & .055 & .767 & .626 & .842 & .793 & .850 & .703 & .689 & .719 & .077 \\
       SCRN$_{19}$  & .789 & .575 & .789 & .817 & .880 & .593 & .651 & .699 & .047 & .830 & .698 & .864 & .854 & .897 & .744 & .757 & .793 & .059 \\
       F$^3$Net$_{20}$  & .739 & .544 & .818 & .795 & .819 & .588 & .593 & .609 & .051 & .780 & .656 & .853 & .824 & .848 & .710 & .705 & .719 & .070 \\
       CSNet$_{20}$  & .778 & .569 & .791 & .810 & .871 & .589 & .635 & .679 & .047 & .750 & .603 & .812 & .773 & .793 & .672 & .655 & .669 & .088 \\
       SSAL$_{20}$  & .668 & .454 & .782 & .768 & .789 & .529 & .527 & .535 & .066 & .699 & .561 & .805 & .780 & .812 & .653 & .644 & .654 & .093 \\
       ITSD$_{20}$   & .767 & .557 & .787 & .808 & .861 & .573 & .615 & .658 & .051 & .811 & .680 & .855 & .845 & .883 & .717 & .729 & .762 & .064 \\
       UCNet$_{20}$   & .776 & .633 & .867 & .857 & .867 & .673 & .681 & .691 & .042 & .811 & .729 & .883 & .871 & .886 & .776 & .775 & .782 & .055 \\
       VST$_{21}$   & .781 & .604 & .837 & .837 & .877 & .620 & .653 & .682 & .042 & .831 & .732 & .887 & .877 & .901 & .758 & .771 & .792 & .050 \\
		\midrule
   \multicolumn{19}{c}{\textbf{Camouflaged Object Detection}} \\
   \midrule
        SINet$_{20}$     & .771 & .551 & .797 & .806 & .868 & .593 & .634 & .676 & .051 & .808 & .723 & .882 & .871 & .883 & .768 & .769 & .775 & .058 \\
        SLSR$_{21}$      & .804 & .673 & .882 & .880 & .892 & .699 & .715 & .732 & .037 & .840 & .766 & .902 & .895 & .907 & .802 & .804 & .815 & .048 \\
        PFNet$_{21}$     & .800 & .660 & .868 & .877 & .890 & .676 & .701 & .725 & .040 & .829 & .745 & .892 & .888 & .898 & .779 & .784 & .799 & .053 \\
        MGL-R$_{21}$     & .814 & .666 & .865 & .852 & .890 & .681 & .711 & .738 & .035 & .833 & .740 & .889 & .867 & .893 & .778 & .782 & .800 & .052 \\
        UJSC$_{21}$      & .809 & .684 & .882 & .884 & .891 & .705 & .721 & .738 & .035 & .842 & .771 & .903 & .898 & .907 & .803 & .806 & .816 & .047 \\
        C$^2$FNet$_{21}$    & .813 & .686 & .886 & .890 & .900 & .703 & .723 & .743 & .036 & .838 & .762 & .898 & .897 & .904 & .788 & .795 & .810 & .049 \\
        UGTR$_{21}$     & .817 & .666 & .850 & .853 & .890 & .671 & .712 & .741 & .036 & .839 & .747 & .886 & .875 & .899 & .778 & .787 & .807 & .052 \\
        BSA-Net$_{22}$  & .818 & .699 & .894 & .891 & .901 & .723 & .738 & .753 & .034 & .841 & .771 & .903 & .897 & .907 & .805 & .808 & .817 & .048 \\
        OCE-Net$_{22}$  & .827 & .707 & .883 & .894 & .905 & .718 & .741 & .764 & .033 & \underline{.853} & .785 & .904 & .903 & .913 & .812 & .818 & .831 & .045 \\
        BGNet$_{22}$    & .831 & .722 & \textbf{.902} & \textbf{.901} & \underline{.911} & \underline{.739} & .753 & .774 & .033 & .851 & \underline{.788} & \underline{.911} & \underline{.907} & \underline{.916} & .813 & \underline{.820} & \underline{.833} & .044 \\
        SegMaR$_{22}$   & .833 & .724 & .893 & \underline{.899} & .906 & \underline{.739} & .757 & .774 & .034 & --   & --     & --     & --     & --     & --     & --     & --     & --     \\
        ZoomNet$_{22}$  & \underline{.838} & \underline{.729} & .892 & .888 & \underline{.911} & \textbf{.741} & \underline{.766} & \underline{.780} & \underline{.029} & \underline{.853} & .784 & .904 & .896 & .912 & \underline{.814} & .818 & .828 & \underline{.043} \\
        SINet-v2$_{22}$ & .815 & .680 & .863 & .887 & .906 & .682 & .718 & .752 & .037 & .847 & .770 & .898 & .903 & .914 & .792 & .805 & .823 & .048 \\

   \midrule
   Ours & \textbf{.851} & \textbf{.735} & \underline{.900} & .895 & \textbf{.930} & .736 & \textbf{.769} & \textbf{.794} & \textbf{.026} & \textbf{.879} & \textbf{.816} & \textbf{.923} & \textbf{.915} & \textbf{.937} & \textbf{.826} & \textbf{.843} & \textbf{.859} & \textbf{.035} \\
   \bottomrule
	\end{tabular}
	}
\end{table*}

\begin{table*}[]
	\centering
	\renewcommand{\tabcolsep}{1.2mm}
	\caption{$S_m$ results for each sub-class on COD10K \cite{fan2020camouflaged}. \textbf{Bolded} and \underline{underlined} denote the best and second best scores.} 
	\label{tab:subclass}
	\scalebox{0.73}{
	\begin{tabular}{l|cccccccccccc}
	\toprule
 \textit{\textbf{Sub-class}} & \textbf{ITSD}  & \textbf{UCNet} & \textbf{SINet} & \textbf{SLSR}  & \textbf{PFNet} & \textbf{MGL-R} & \textbf{UJSC}  & \textbf{C$^2$FNet} & \textbf{OCE-Net} & \textbf{ZoomNet}   & \textbf{SINet-v2}   & \textbf{Ours}   \\
\midrule
Amphibian-Frog  & 0.761 & 0.790 & 0.785 & 0.814  & 0.815  & 0.813  & 0.794  & 0.820  & 0.809  & \textbf{0.855} & 0.837  & \underline{0.852} \\
\rowcolor{gray!25}Amphibian-Toad  & 0.836 & 0.847 & 0.850 & 0.863  & 0.866  & 0.877  & 0.867  & 0.866  & 0.888  & \underline{ 0.885} & 0.870  & \textbf{0.893} \\
Aquatic-BatFish & 0.813 & 0.820 & 0.836 & 0.809  & 0.867  & 0.786  & 0.817  & 0.834  & 0.906  & \underline{ 0.890} & 0.873  & \textbf{0.907} \\
\rowcolor{gray!25}Aquatic-ClownFish  & 0.692 & 0.707 & 0.693 & 0.784  & 0.805  & 0.608  & 0.709  & 0.737  & 0.771  & \underline{ 0.813} & 0.787  & \textbf{0.851} \\
Aquatic-Crab  & 0.791 & 0.788 & 0.804 & 0.817  & 0.805  & \underline{ 0.839} & 0.811  & 0.828  & \underline{ 0.839} & 0.836  & 0.815  & \textbf{0.864} \\
\rowcolor{gray!25}Aquatic-Crocodile  & 0.752 & 0.815 & 0.761 & 0.783  & 0.753  & 0.785  & 0.808  & 0.817  & \underline{ 0.843} & 0.829  & 0.825  & \textbf{0.857} \\
Aquatic-CrocodileFish   & 0.689 & 0.785 & 0.734 & 0.791  & 0.780  & \underline{ 0.815} & 0.751  & 0.764  & 0.738  & 0.805  & 0.746  & \textbf{0.846} \\
\rowcolor{gray!25}Aquatic-Fish  & 0.780 & 0.791 & 0.767 & 0.816  & 0.799  & 0.821  & 0.821  & 0.818  & 0.835  & \underline{ 0.841} & 0.834  & \textbf{0.854} \\
Aquatic-Flounder   & 0.814 & 0.812 & 0.786 & 0.857  & 0.850  & 0.872  & 0.873  & 0.857  & \underline{ 0.895} & 0.880  & 0.889  & \textbf{0.922} \\
\rowcolor{gray!25}Aquatic-FrogFish   & 0.831 & 0.849 & 0.748 & 0.848  & 0.840  & 0.844  & \underline{ 0.894} & 0.868  & 0.879  & \textbf{0.925} & \underline{ 0.894} & \textbf{0.925} \\
Aquatic-GhostPipefish   & 0.794 & 0.774 & 0.779 & 0.821  & 0.819  & 0.832  & 0.823  & 0.831  & 0.840  & \underline{ 0.849} & 0.817  & \textbf{0.872} \\
\rowcolor{gray!25}Aquatic-LeafySeaDragon  & 0.587 & 0.609 & 0.576 & 0.654  & 0.625  & \underline{ 0.714} & 0.641  & 0.633  & 0.658  & 0.691  & 0.670  & \textbf{0.782} \\
Aquatic-Octopus & 0.846 & 0.827 & 0.843 & 0.873  & 0.863  & 0.895  & 0.865  & 0.869  & \underline{ 0.897} & \textbf{0.889} & 0.887  & 0.885  \\
\rowcolor{gray!25}Aquatic-Pagurian   & 0.654 & 0.624 & 0.643 & 0.682  & 0.640  & 0.683  & 0.679  & 0.650  & 0.646  & \textbf{0.724} & 0.698  & \underline{ 0.710} \\
Aquatic-Pipefish   & 0.718 & 0.755 & 0.726 & 0.782  & 0.774  & 0.805  & 0.780  & 0.792  & \underline{ 0.810} & 0.807  & 0.781  & \textbf{0.828} \\
\rowcolor{gray!25}Aquatic-ScorpionFish & 0.779 & 0.731 & 0.740 & 0.799  & 0.773  & 0.753  & 0.806  & 0.778  & 0.815  & \underline{ 0.834} & 0.808  & \textbf{0.851} \\
Aquatic-SeaHorse   & 0.793 & 0.789 & 0.799 & 0.823  & 0.798  & 0.814  & 0.826  & 0.824  & \underline{ 0.835} & 0.823  & 0.823  & \textbf{0.851} \\
\rowcolor{gray!25}Aquatic-Shrimp  & 0.670 & 0.656 & 0.718 & 0.700  & 0.737  & 0.730  & 0.714  & 0.741  & 0.731  & \underline{ 0.787} & 0.735  & \textbf{0.819} \\
Aquatic-Slug  & 0.701 & 0.786 & 0.792 & 0.594  & \textbf{0.836} & 0.770  & 0.743  & \underline{ 0.803} & 0.624  & 0.776  & 0.729  & 0.696  \\
\rowcolor{gray!25}Aquatic-StarFish   & 0.797 & 0.876 & 0.799 & 0.856  & 0.852  & 0.846  & \textbf{0.903} & \underline{ 0.892} & 0.868  & \underline{ 0.892} & 0.890  & 0.889  \\
Aquatic-Stingaree  & 0.779 & 0.669 & 0.724 & 0.789  & 0.785  & 0.747  & 0.781  & 0.791  & 0.815  & \underline{ 0.818} & 0.815  & \textbf{0.881} \\
\rowcolor{gray!25}Aquatic-Turtle  & 0.804 & 0.803 & 0.774 & 0.823  & 0.838  & 0.814  & 0.822  & 0.785  & 0.833  & \textbf{0.898} & 0.760  & \underline{ 0.883} \\
Flying-Bat & 0.789 & 0.740 & 0.769 & 0.782  & 0.838  & 0.836  & 0.795  & 0.817  & 0.844  & 0.822  & \underline{ 0.847} & \textbf{0.875} \\
\rowcolor{gray!25}Flying-Bee & 0.743 & 0.685 & 0.727 & 0.786  & 0.727  & 0.749  & 0.741  & \underline{ 0.774} & 0.783  & 0.743  & \textbf{0.777} & 0.680  \\
Flying-Beetle   & 0.916 & 0.917 & 0.911 & \underline{ 0.931} & 0.903  & 0.821  & 0.923  & 0.922  & 0.926  & \underline{ 0.931} & 0.903  & \textbf{0.932} \\
\rowcolor{gray!25}Flying-Bird   & 0.796 & 0.815 & 0.807 & 0.830  & 0.826  & 0.841  & 0.836  & 0.846  & 0.851  & \underline{ 0.867} & 0.835  & \textbf{0.873} \\
Flying-Bittern  & 0.844 & 0.848 & 0.844 & 0.860  & 0.855  & \underline{ 0.867} & \underline{ 0.867} & 0.840  & 0.863  & \textbf{0.895} & 0.849  & 0.865  \\
\rowcolor{gray!25}Flying-Butterfly   & 0.823 & 0.856 & 0.828 & 0.871  & 0.862  & 0.864  & 0.868  & 0.881  & 0.878  & 0.882  & \underline{ 0.883} & \textbf{0.885} \\
Flying-Cicada   & 0.834 & 0.843 & 0.851 & 0.875  & 0.887  & 0.891  & 0.886  & 0.875  & 0.900  & \textbf{0.916} & 0.883  & \underline{ 0.909} \\
\rowcolor{gray!25}Flying-Dragonfly   & 0.790 & 0.785 & 0.772 & 0.825  & 0.820  & 0.845  & 0.828  & 0.828  & 0.823  & \underline{ 0.840} & 0.837  & \textbf{0.886} \\
Flying-Frogmouth   & 0.929 & 0.932 & 0.896 & 0.941  & 0.945  & 0.936  & 0.932  & 0.936  & 0.942  & \textbf{0.961} & 0.941  & \underline{ 0.947} \\
\rowcolor{gray!25}Flying-Grasshopper & 0.797 & 0.808 & 0.801 & 0.823  & 0.809  & 0.835  & 0.831  & 0.830  & 0.852  & \underline{ 0.853} & 0.833  & \textbf{0.874} \\
Flying-Heron  & 0.797 & 0.795 & 0.780 & 0.845  & 0.832  & 0.827  & 0.840  & 0.818  & 0.840  & \textbf{0.889} & 0.823  & \underline{ 0.866} \\
\rowcolor{gray!25}Flying-Katydid  & 0.760 & 0.768 & 0.771 & 0.797  & 0.798  & 0.833  & 0.811  & 0.824  & 0.840  & \underline{ 0.847} & 0.809  & \textbf{0.853} \\
Flying-Mantis   & 0.733 & 0.750 & 0.721 & 0.765  & 0.755  & 0.767  & 0.780  & 0.767  & 0.771  & \underline{ 0.804} & 0.775  & \textbf{0.835} \\
\rowcolor{gray!25}Flying-Mockingbird & 0.767 & 0.795 & 0.735 & 0.834  & 0.800  & 0.796  & 0.832  & 0.843  & 0.821  & \underline{ 0.863} & 0.838  & \textbf{0.875} \\
Flying-Moth   & 0.838 & 0.890 & 0.863 & 0.894  & 0.893  & 0.877  & 0.908  & 0.901  & 0.929  & 0.916  & \underline{ 0.917} & \textbf{0.941} \\
\rowcolor{gray!25}Flying-Owl & 0.805 & 0.834 & 0.832 & 0.861  & 0.855  & 0.865  & 0.851  & 0.868  & \underline{ 0.890} & \textbf{0.896} & 0.868  & 0.876  \\
Flying-Owlfly   & 0.837 & 0.782 & 0.772 & 0.851  & 0.752  & 0.845  & 0.866  & 0.861  & \textbf{0.886} & \underline{ 0.879} & 0.863  & 0.872  \\
\rowcolor{gray!25}Other-Other   & 0.753 & 0.755 & 0.737 & 0.807  & 0.764  & 0.812  & 0.809  & 0.790  & 0.812  & \underline{ 0.846} & 0.779  & \textbf{0.899} \\
Terrestrial-Ant & 0.644 & 0.644 & 0.635 & 0.705  & 0.659  & 0.708  & 0.643  & 0.658  & 0.679  & \underline{ 0.742} & 0.669  & \textbf{0.743} \\
\rowcolor{gray!25}Terrestrial-Bug & 0.845 & 0.820 & 0.844 & 0.828  & 0.840  & 0.872  & 0.843  & 0.848  & 0.897  & \underline{ 0.865} & 0.856  & \textbf{0.874} \\
Terrestrial-Cat & 0.712 & 0.698 & 0.712 & 0.746  & 0.751  & 0.763  & 0.745  & 0.768  & 0.772  & \underline{ 0.785} & 0.772  & \textbf{0.827} \\
\rowcolor{gray!25}Terrestrial-Caterpillar & 0.686 & 0.737 & 0.704 & 0.756  & 0.745  & 0.753  & 0.786  & 0.746  & 0.766  & \underline{ 0.794} & 0.776  & \textbf{0.813} \\
Terrestrial-Centipede   & 0.707 & 0.697 & 0.677 & 0.744  & 0.682  & 0.746  & 0.645  & 0.731  & 0.767  & 0.733  & \underline{ 0.762} & \textbf{0.791} \\
\rowcolor{gray!25}Terrestrial-Chameleon   & 0.765 & 0.746 & 0.759 & 0.786  & 0.776  & \underline{ 0.834} & 0.814  & 0.810  & 0.803  & 0.833  & 0.804  & \textbf{0.845} \\
Terrestrial-Cheetah  & 0.780 & 0.762 & 0.786 & 0.813  & 0.794  & 0.821  & \underline{ 0.832} & \underline{ 0.832} & 0.828  & 0.821  & 0.826  & \textbf{0.851} \\
\rowcolor{gray!25}Terrestrial-Deer   & 0.691 & 0.711 & 0.701 & 0.738  & 0.737  & 0.751  & 0.761  & 0.760  & 0.757  & \underline{ 0.791} & 0.757  & \textbf{0.798} \\
Terrestrial-Dog & 0.667 & 0.665 & 0.669 & 0.677  & 0.690  & 0.706  & 0.677  & 0.690  & 0.712  & \underline{ 0.729} & 0.707  & \textbf{0.786} \\
\rowcolor{gray!25}Terrestrial-Duck   & 0.694 & 0.694 & 0.693 & 0.709  & \textbf{0.800} & 0.732  & 0.710  & 0.747  & 0.750  & 0.726  & 0.746  & \underline{ 0.784} \\
Terrestrial-Gecko  & 0.789 & 0.831 & 0.825 & 0.861  & 0.830  & 0.815  & \underline{ 0.868} & 0.867  & 0.879  & 0.856  & 0.848  & \textbf{0.908} \\
\rowcolor{gray!25}Terrestrial-Giraffe  & 0.747 & 0.786 & 0.773 & 0.809  & 0.821  & 0.799  & 0.788  & 0.779  & 0.808  & \underline{ 0.826} & 0.784  & \textbf{0.846} \\
Terrestrial-Grouse & 0.919 & 0.918 & 0.927 & 0.918  & 0.938  & 0.936  & 0.934  & 0.925  & 0.948  & \underline{ 0.941} & 0.921  & \textbf{0.942} \\
\rowcolor{gray!25}Terrestrial-Human  & 0.753 & 0.712 & 0.742 & 0.790  & 0.783  & 0.765  & 0.768  & 0.792  & \textbf{0.827} & 0.781  & \underline{ 0.817} & 0.797  \\
Terrestrial-Kangaroo & 0.762 & 0.772 & 0.737 & 0.748  & 0.761  & \underline{ 0.815} & 0.788  & 0.806  & 0.729  & 0.800  & \textbf{0.816} & 0.802  \\
\rowcolor{gray!25}Terrestrial-Leopard  & 0.808 & 0.784 & 0.805 & 0.798  & 0.836  & \underline{ 0.847} & 0.800  & 0.834  & 0.826  & 0.846  & 0.823  & \textbf{0.851} \\
Terrestrial-Lion   & 0.773 & 0.807 & 0.761 & 0.832  & 0.818  & 0.814  & 0.815  & 0.833  & \underline{ 0.843} & 0.814  & 0.813  & \textbf{0.859} \\
\rowcolor{gray!25}Terrestrial-Lizard & 0.786 & 0.804 & 0.800 & 0.820  & 0.819  & 0.829  & 0.830  & 0.823  & 0.838  & \underline{ 0.852} & 0.830  & \textbf{0.853} \\
Terrestrial-Monkey & 0.829 & 0.644 & 0.675 & 0.808  & 0.720  & 0.855  & 0.877  & 0.835  & 0.797  & \underline{ 0.898} & 0.888  & \textbf{0.913} \\
\rowcolor{gray!25}Terrestrial-Rabbit & 0.827 & 0.814 & 0.804 & 0.838  & 0.841  & 0.840  & 0.841  & 0.840  & 0.852  & \underline{ 0.854} & 0.843  & \textbf{0.887} \\
Terrestrial-Reccoon  & 0.756 & 0.790 & 0.738 & 0.748  & 0.774  & 0.619  & 0.780  & 0.788  & 0.801  & \textbf{0.837} & 0.766  & \underline{ 0.791} \\
\rowcolor{gray!25}Terrestrial-Sciuridae   & 0.804 & 0.811 & 0.821 & 0.798  & 0.831  & 0.841  & 0.838  & 0.852  & 0.844  & \textbf{0.897} & 0.842  & \underline{ 0.856} \\
Terrestrial-Sheep  & 0.487 & 0.754 & 0.490 & 0.750  & 0.582  & 0.565  & 0.561  & \underline{ 0.686} & 0.540  & \textbf{0.761} & 0.500  & 0.493  \\
\rowcolor{gray!25}Terrestrial-Snake  & 0.776 & 0.796 & 0.796 & 0.816  & 0.824  & 0.835  & 0.819  & 0.839  & 0.816  & \textbf{0.862} & 0.831  & \underline{ 0.854} \\
Terrestrial-Spider & 0.716 & 0.724 & 0.736 & 0.757  & 0.759  & 0.782  & 0.766  & 0.769  & 0.793  & \underline{ 0.802} & 0.771  & \textbf{0.808} \\
\rowcolor{gray!25}Terrestrial-StickInsect & 0.667 & 0.701 & 0.658 & 0.746  & 0.701  & 0.695  & 0.726  & 0.730  & 0.757  & \underline{ 0.753} & 0.696  & \textbf{0.762} \\
Terrestrial-Tiger  & 0.637 & 0.646 & 0.662 & 0.663  & 0.675  & \underline{ 0.706} & 0.669  & 0.695  & 0.689  & 0.700  & 0.703  & \textbf{0.734} \\
\rowcolor{gray!25}Terrestrial-Wolf   & 0.731 & 0.731 & 0.737 & 0.707  & 0.730  & 0.747  & 0.754  & 0.747  & \underline{ 0.757} & \textbf{0.792} & 0.749  & 0.749  \\
Terrestrial-Worm   & 0.659 & 0.710 & 0.682 & 0.739  & 0.792  & 0.766  & 0.804  & 0.691  & 0.782  & \underline{ 0.808} & 0.806  & \textbf{0.812}\\

\bottomrule
\end{tabular}
	}
\end{table*}

\begin{figure*}[thp!]
	\centering
	\begin{overpic}[width=0.9\linewidth]{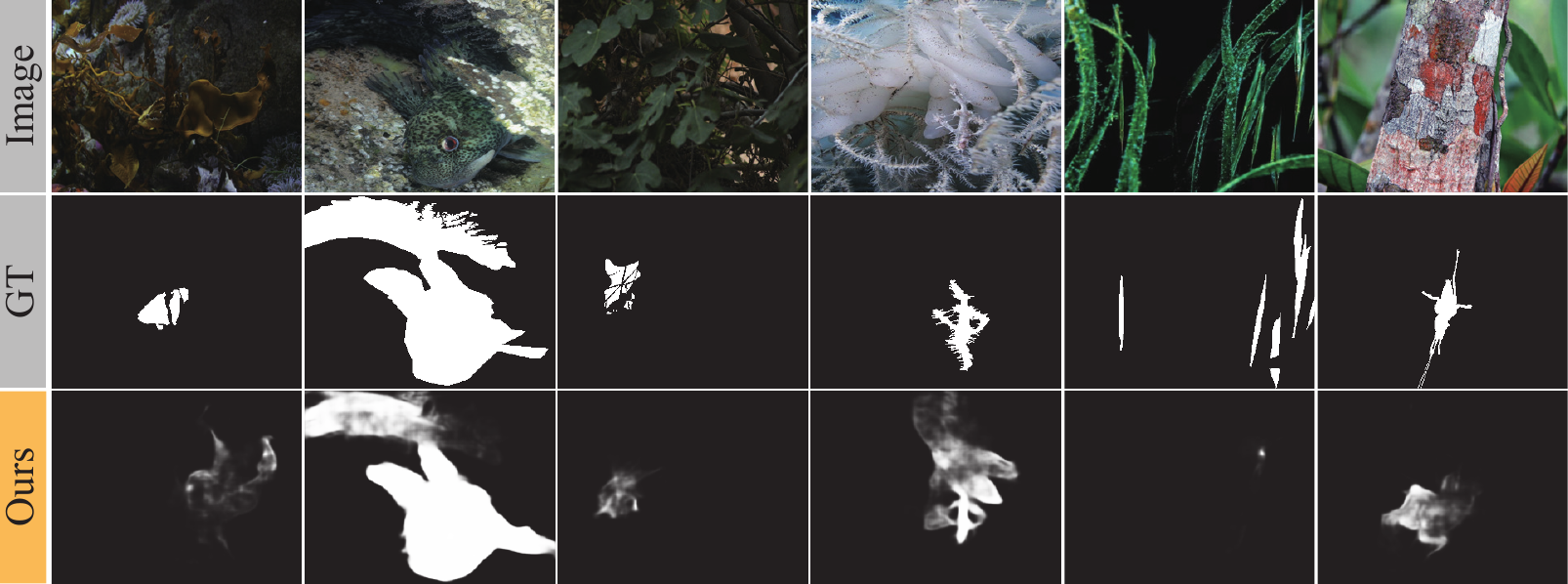}
	\end{overpic}
	\caption{Failure cases in very challenging scenarios. Please zoom in for details.
	}
	\label{fig:VC-failure-cases}
\end{figure*}

\subsection{More Ablation Experiments on FSD} 
\cref{tab:ablation_FSD} shows more ablation experiments of feature shrinkage decoder on COD10K and NC4K datasets. {\ding{172}} is similar to \cite{ma2021pyramidal}, which adjusts our decoder to pairwise feature aggregation with overlapping, and removes lateral supervision and feature interaction within the same layer. {\ding{173}} is a decoder that adds the lateral supervision and feature interaction within the same layer to {\ding{172}}. Our method differs from {\ding{173}} in that our method is pairwise feature aggregation without overlapping. Note that we retain other modules in these experiments. 

By comparison of {\ding{172}} and ours, we can see that our decoder achieves superior performance over {\ding{172}}. In particular, our decoder significantly improves performance by 1.5\%, 1.3\%, 1.6\% and 1.1\% for $F^w_{\beta}$, $F^m_{\beta}$, $E^m_{\phi}$ and $E^x_{\phi}$, respectively, on the NC4K dataset. 
Although {\ding{172}} and ours both progressively aggregate adjacent features through a layer-by-layer shrinkage pyramid to accumulate features for object prediction, our decoder introduces lateral supervision and feature flow within the same layer, which forces the decoder to accumulate more critical camouflaged object cues for better object segmentation. 

By comparison of {\ding{173}} and ours, we reduce the aggregation operations (\textit{i.e.}, AIM) to alleviate the decoder structure by fusing adjacent features without overlapping. Specifically, {\ding{173}} contains 11 layers and 66 AIMs, while our decoder only contains 4 layers and 12 AIMs, which greatly reduces the computation. However, our decoder still achieves slightly better performance than {\ding{173}} with fewer aggregation operations. 
Experiments demonstrate the superior performance of the proposed FSD to other decoder structures.

\subsection{Failure Cases}
Although our proposed model achieves state-of-the-art performance, it does not detect camouflaged objects well in some very challenging scenes. As shown in \cref{fig:VC-failure-cases}, the results in the first and last three columns indicate the difficulty of detecting camouflaged objects under very low lighting conditions and a very similar appearance to the background, respectively, which are potential directions for improvement in our future work.

\end{document}